\documentclass[journal,a4paper]{IEEEtran}

\newif\ifanonymous

\anonymousfalse       %

\newif\ifieeemembers
\ieeemembersfalse       %

\usepackage{cite}
\usepackage{placeins}
\usepackage{url}
\usepackage{amsmath}
\usepackage{amssymb}
\usepackage{graphicx}
\usepackage{tabularx}
\usepackage{booktabs}
\usepackage{bm}
\usepackage{multirow}
\usepackage{siunitx}
\usepackage{tikz}

\usepackage[caption=false,font=footnotesize]{subfig}

\usepackage{amsfonts}
\usepackage[bottom]{footmisc}
\usepackage{xcolor}
\usepackage{mathtools}
\usepackage{tikz-3dplot}
\usepackage{pdfpages}
\usepackage{tikzpagenodes}
\usepackage{layout}
\usepackage{makecell}
\usepackage{accents}

\usepackage{enumitem}

\usetikzlibrary{
  backgrounds,
  positioning,
  calc,
  arrows.meta,
  shapes,
  shapes.misc,
  shapes.symbols,
  patterns,
  fit,
  decorations.pathreplacing,
  quotes,
  3d
}

\usepackage{cnnbox}

\usepackage[hidelinks]{hyperref}
\usepackage{cleveref}

\newcommand{\authmember}[2]{%
  \ifieeemembers,~\IEEEmembership{#1,~IEEE#2}\else#2\fi}

\definecolor{nordOrigin} {HTML}{2E3440}
\definecolor{nordDeep}   {HTML}{3B4252}
\definecolor{nordDim}    {HTML}{434C5E}
\definecolor{nordPolar}  {HTML}{4C566A}
\definecolor{nordMist}   {HTML}{D8DEE9}
\definecolor{nordSnow}   {HTML}{E5E9F0}
\definecolor{nordArc}    {HTML}{ECEFF4}
\definecolor{nordTeal}   {HTML}{8FBCBB}
\definecolor{nordFrost}  {HTML}{88C0D0}
\definecolor{nordIce}    {HTML}{81A1C1}
\definecolor{nordBlue}   {HTML}{5E81AC}
\definecolor{nordRed}    {HTML}{BF616A}
\definecolor{nordOrange} {HTML}{D08770}
\definecolor{nordYellow} {HTML}{EBCB8B}
\definecolor{nordGreen}  {HTML}{A3BE8C}
\definecolor{nordPurple} {HTML}{B48EAD}

\crefname{figure}{Fig.}{Figs.}          \Crefname{figure}{Figure}{Figures}
\crefname{subfigure}{Fig.}{Figs.}       \Crefname{subfigure}{Figure}{Figures}
\crefname{table}{Tab.}{Tabs.}           \Crefname{table}{Table}{Tables}
\crefname{section}{Sec.}{Secs.}         \Crefname{section}{Section}{Sections}
\crefname{subsection}{Sec.}{Secs.}      \Crefname{subsection}{Section}{Sections}
\crefname{equation}{Eq.}{Eqs.}
\Crefname{equation}{Equation}{Equations}
\crefname{assumption}{Assumption}{Assumptions}
\Crefname{assumption}{Assumption}{Assumptions}
\crefname{algorithm}{Alg.}{Algs.}
\Crefname{algorithm}{Algorithm}{Algorithms}
\AtBeginDocument{}

\makeatletter
\let\IEEEorig@makecaption\@makecaption
\long\def\@makecaption#1#2{%
  \ifx\@captype\@IEEEtablestring
    \footnotesize\bgroup\par\parindent=0pt\@IEEEtabletopskipstrut
    {\normalfont\footnotesize #1.}\nobreakspace\nobreakspace #2\par
    \addvspace{0.5\baselineskip}\egroup
    \@IEEEtablecaptionsepspace
  \else
    \IEEEorig@makecaption{#1}{#2}%
  \fi}
\makeatother

\def\diag{\operatorname{diag}}
\newcommand{\skewm}[1]{[#1]_\times}   %
\newcommand{\veemap}[1]{[#1]^\vee}    %

\newcommand{\uprightsubscript}[1]{_{\textnormal{#1}}}
\begingroup\lccode`~ =`?\lowercase{\endgroup\let~}\uprightsubscript
\AtBeginDocument{\mathcode`?="8000 }
\newcommand{\uprightsuperscript}[1]{^{\textnormal{#1}}}
\begingroup\lccode`~=`*\lowercase{\endgroup\let~}\uprightsuperscript
\AtBeginDocument{\mathcode`*="8000 }

\newtheorem{assumption}{Assumption}

\tikzset{
  figblock/.style={draw, thick, rounded corners=3pt, align=center, inner
  sep=4pt},
  fighue/.style={fill=#1!20, draw=#1!70!black},
  figgroup/.style={draw=nordPolar, densely dashed},
  figline/.style={-{Stealth[length=6pt]}, line width=1pt, draw=#1!70!black},
  figaxis/.style={-{Stealth[length=4pt]}, line width=1.6pt, draw=#1},
  figlead/.style={line width=0.6pt, draw=#1!70!black},
  figlabel/.style={font=\small, inner sep=2pt, text=#1!70!black},
  figplate/.style={fill=white, fill opacity=0.75, text opacity=1,
    draw=#1!70!black, line width=0.6pt, rounded corners=3pt,
    inner sep=2pt},
  figlegend/.style={draw=black, fill=white, inner xsep=6pt, inner ysep=5pt,
    font=\small},
}

\def\cnnStem{nordOrange}
\def\cnnStage{nordPurple}
\def\cnnPool{nordMist}
\def\cnnAdaptPool{nordFrost}
\def\cnnFlatten{nordGreen}
\def\cnnMlp{nordYellow}
\def\cnnOut{nordIce}

\newcommand{\tabstyle}{\setlength{\tabcolsep}{6pt}\renewcommand{\arraystretch}{1.15}}
\newcommand{\tabnote}[1]{\\[3pt]{\footnotesize #1}}
\newcolumntype{L}[1]{>{\setlength\hsize{#1\hsize}\raggedright\arraybackslash}X}
\newcolumntype{C}[1]{>{\setlength\hsize{#1\hsize}\centering\arraybackslash}X}

\newcommand{\datasetstatement}{%
  \ifanonymous
    The dataset will be released upon publication.
  \else
    The dataset is available at \url{https://doi.org/10.5281/zenodo.23209032}.
  \fi
}

\newcommand{\videostatement}{%
  \ifanonymous
    An anonymized video accompanies the submission.
  \else
    A video of the experiments is available at
    \url{https://youtube.com/playlist?list=PLejKMZmW8AvI}.
  \fi
}

\newcommand{\additionaltrialvideostatement}{%
  \ifanonymous
    Footage of these trials is included in the anonymized video accompanying
    the submission.
  \else
    Footage of these trials is included in the video referenced above.
  \fi
}

\newcommand{\identifiedonly}[1]{%
  \ifanonymous
  \else
    #1%
  \fi
}

\newcommand{\anonymousonly}[1]{%
  \ifanonymous
    #1%
  \fi
}

\title{Tactile Reconstruction of Contact Task Frames and Forces for Hybrid
  Force/Motion Control}

\ifanonymous

  \author{Anonymous Authors}

\else

  \author{%
    Antonio~Rapuano$^1$\authmember{Student Member}{,}
    Simone~Orelli$^1$\authmember{Student Member}{,}
    Barbara~Bazzana$^2$\authmember{Member}{,}
    Antonio~Franchi$^{1,2}$\authmember{Fellow}{,}
    and~Alessandro~De~Luca$^1$\authmember{Life Fellow}{}%
    \thanks{A. Rapuano and S. Orelli contributed equally to this work.}%
    \thanks{%
      $^{1}$Department of Computer, Control and Management Engineering,
      Sapienza University of Rome, 00185 Rome, Italy.
      {\tt\footnotesize schol@r-rapuano.com},
      {\tt\footnotesize schol@r-orelli.eu},
      {\tt\footnotesize schol@r-franchi.eu},
      {\tt\footnotesize deluca@diag.uniroma1.it}.%
    }%
    \thanks{%
      $^{2}$Robotics and Mechatronics Department, Faculty of Electrical
      Engineering, Mathematics and Computer Science, University of Twente,
      7500 AE Enschede, The Netherlands.
      {\tt\footnotesize b.bazzana@utwente.nl},
      {\tt\footnotesize schol@r-franchi.eu}.%
    }%
    \thanks{%
      This work has been partly supported by the European Union under Grant
      AUTOASSESS 101120732, and the 3rd Technology Exchange Programme
      (SOAR-TOUCH) of the European Union Grant euROBIN 101070596.%
    }%
  }

\fi

\begin{document}

\maketitle

\begin{abstract}
  Hybrid force/motion control requires knowledge of the interaction force and of
  a task frame defining the force- and motion-controlled directions.
  These quantities are usually obtained from force/torque sensing or model-based
  residuals, often assuming also a nominal environment model.
  This work addresses the online estimation of the contact force and a possibly
  time-varying task frame using only soft optical tactile sensing, under the
  assumption of locally planar contact with a negligible contact moment.
  The proposed method maps a single image of the deformed elastomer of a soft
  optical tactile sensor to observable contact variables: indentation depth, two
  surface-to-sensor tilt angles, and 3D contact force, each with a per-sample
  uncertainty estimate.
  The mapping is learned through a self-labeling acquisition procedure, in which
  a manipulator imposes controlled contacts while an auxiliary Force/Torque
  sensor is used offline to provide ground-truth labels.
  The tactile measurement is then fused with robot proprioceptive data in an
  Extended Kalman Filter, producing a continuously updated estimate of the
  contact task frame and of the interaction force.
  Control experiments with a DigiTac sensor mounted on a UR10 manipulator
  demonstrate closed-loop contact force regulation against a flat rigid board in
  linear and angular motion by a human operator, with touch as the only
  exteroceptive feedback.
\end{abstract}


\section{Introduction}
\label{sec:intro}

\looseness=-1 \IEEEPARstart{P}{hysical} interaction with the environment is
central to many robotic applications, including manufacturing, inspection,
service robotics, and human-robot collaboration.
When the task specifies both how the robot should move and how it should
exchange forces with the environment, standard control tools include admittance,
impedance, and hybrid force/motion control.
Admittance control reacts to a measured contact force with a velocity
command~\cite{1977-Whit}.
Impedance control addresses physical interaction by shaping the dynamic
relationship between motion and force~\cite{1985-Hog}.
Hybrid force/motion control assigns force along selected task directions and
motion along the complementary ones, as formalized in the classical task-frame
formulation~\cite{1981-RaiCra}.
These approaches and their relation to manipulator interaction control are
reviewed in~\cite{2025-SicVilOriDeL}.

\input{assets/motivation}

\looseness=-1 Hybrid force/motion control requires, in addition to robot
proprioception, knowledge of the interaction force and of a task frame that
defines the local force- and motion-controlled directions.
In most implementations, this information is obtained from sources other than
local tactile deformation.
Force/Torque (F/T) sensors provide accurate wrench measurements, but they add
dedicated hardware and return a spatially aggregated signal, which is usually
displaced from the actual contact.

When such sensors are unavailable, external wrenches can be estimated from
proprioception through model-based residuals.
Residual-based contact and collision estimation is treated systematically
in~\cite{2017-HadDeLAlb}.
Such estimates have also been used to implement sensorless collision detection
and hybrid force/motion control~\cite{2005-DeLMat}.
Their integration into hybrid force/velocity control has been demonstrated for
physical human-robot collaboration tasks~\cite{2016-MagDeL}.
However, residual methods estimate generalized forces acting on the robot
structure.
Recovering the local contact point, surface direction, and task frame from these
estimates requires additional assumptions and remains sensitive to modeling
errors~\cite{2014-MagFlaDeL}.

The task frame is often specified from a geometric model of the
environment~\cite{1996-BruDeS}.
When the contact geometry is not known accurately, several approaches instead
estimate its relevant directions during interaction.
Dynamic hybrid position/force control estimating unknown environmental
constraints was achieved in~\cite{1993-YosSud}.
Force/position control on an unknown planar surface was addressed
in~\cite{2009-KarDou} by adapting the estimated surface orientation using force
and robot motion measurements.
Contact force and the unknown surface gradient can be jointly estimated using
the latter to construct the force- and motion-control
subspaces~\cite{2020-GutArt}.
The surface normal can be also estimated from F/T measurements and commanded
motion while compensating for friction-induced bias~\cite{2024-NasWan}.
These approaches demonstrate that the task frame can be recovered online.
However, the geometric information is inferred through external force measures,
proprioception, and closed-loop motion signals that are themselves influenced by
the contact mechanics and the resulting interaction.

Tactile sensing offers a complementary route because it observes the contact
patch rather than only the net wrench or robot motion.
Soft skins provide spatially distributed measurements of local contact
deformation~\cite{2010-DahMetValSan}.
Vision-based tactile sensors can use this deformation to estimate both contact
geometry and force~\cite{2017-YuaDonAde}.
Soft Optical Tactile (SOT) sensors encode the deformation of a compliant pad
into an image.
The TacTip family realizes this principle through the motion of internal markers
that follow the deformation of its soft
surface~\cite{2018-WarPesCraWinGiaRosLep}.
The DigiTac sensor adopted in this work combines this biomimetic pin arrangement
with a compact optical design~\cite{2022-LepLinMonLlo}.
The resulting image contains local geometric information about the contact that
is not directly available from a spatially aggregated wrench measurement.

Recent tactile servoing methods exploit this information by learning contact
pose variables from tactile images and regulating them to a reference condition.
Pose servoing using a novel soft optical tactile sensor with learned pose
estimates has been demonstrated in~\cite{2021-LepLlo}.
Pose-and-shear tactile servoing extends this approach by incorporating shear
information into the regulated tactile representation~\cite{2024-LloLep}.
Learned tactile predictions have also been propagated through a Bayesian filter
using an associated estimate of uncertainty~\cite{2021-VazLep}.
However, these tactile servoing approaches typically shape the interaction force
implicitly by regulating a tactile pose or deformation condition, rather than
assigning a contact force within a hybrid force/motion decomposition.

\looseness=-1 Force regression methods are also available that estimate
interaction forces from tactile images.
FeelAnyForce targets 3D contact force estimation across vision-based tactile
sensors~\cite{2025-ShaCadAlaNatFerAlo}.
Learning supervised by finite element analysis has been used to estimate force
distributions from a GelSight Mini sensor~\cite{2025-HelDziFunBelPet}.
Such force estimates do not provide by themselves the local task frame needed to
define force- and motion-controlled directions.
While each item stands by itself, the missing part is the integration of tactile
pose {\em and} force estimation into the contact variables required by hybrid
force/motion control.

This work addresses this gap for locally planar, featureless surfaces with an
applied contact force but with negligible contact moment.
Under these assumptions, different positions on the contact plane as well as
different tilted directions forming the same angle with the normal to the plane
are not observable (indistinguishable) from tactile images.
Nevertheless, these images can encode the observable contact geometry, namely
the indentation depth and the surface-to-sensor tilt, together with the 3D
contact force transmitted through the contact patch.
The proposed method learns a mapping from a single SOT image to this tactile
measurement, with an associated per-sample uncertainty estimate.
\looseness=-1 The tactile measurement is then fused with robot proprioceptive
data to obtain a continuously updated estimate of the contact plane and of the
corresponding contact task frame.
The same estimator tracks the interaction force while contact is active.
These estimates are used by an admittance-based hybrid force/motion controller,
closing all three translational channels of force regulation from tactile
feedback alone.

\looseness=-1 \Cref{fig:motivation} summarizes the resulting architecture.
The auxiliary F/T sensor is excluded from the control loop: it is used only
offline as ground truth to construct training labels and to provide an
independent measurement for experimental comparison.
During operation, the system receives only robot proprioceptive data and tactile
images, reconstructs the contact state, and supplies the resulting estimates of
task frame and interaction force to the controller.

\looseness=-1 To the best of the authors' knowledge, this is the first
integrated system that reconstructs both a contact task frame and a 3D
interaction force from soft optical tactile images, propagates the learned
tactile uncertainty through a state estimator, and uses the reconstructed
quantities to close a hybrid force/motion control loop without F/T feedback or
an environment model.

The specific contributions are the following.
\begin{itemize}[leftmargin=*]

  \item A tactile measurement formulation that retains the contact variables
    that are observable under the adopted locally planar contact model:
    indentation depth, two surface-to-sensor tilt angles, and 3D contact force.
  Using a single image and the regressor, these quantities are estimated
  together with a per-sample uncertainty for Bayesian state estimation.

  \item A self-labeling acquisition framework in which the manipulator generates
    controlled contacts and provides the corresponding geometric labels, using
    an auxiliary F/T sensor for ground truth values.

  \item A tactile-proprioceptive contact state estimator that reconstructs
    online the local contact plane and the task frame while tracking the
    interaction force target during contact.
  The estimated surface geometry is retained, with increasing uncertainty,
  during brief contact breaks.

  \item An experimental demonstration of the complete tactile-driven control
    architecture on a manipulator interacting with a flat rigid board undergoing
    linear and angular motion imposed by a human operator.
  The controller simultaneously regulates the three contact force components,
  the two tangential components to zero and the normal component to a nonzero
  reference, while adapting the sensor orientation to the reconstructed contact
  frame.
  Touch is the only source of exteroceptive feedback used by the control loop.

\end{itemize}

The paper is organized as follows.
\Cref{sec:problem} formulates the problem, stating the contact assumptions and
the force/motion decomposition they induce.
The self-labeling acquisition protocol is described in~\cref{sec:dataset},
while~\cref{sec:network} discusses the tactile measurement regressor.
\Cref{sec:control} presents the contact state estimator and the design of the
hybrid force/motion controller.
The experimental results on training and validation, on contact state
estimation, and on hybrid force/motion control are reported
in~\cref{sec:results}.
Finally, conclusions and future work are summarized in~\cref{sec:conclusion}.
A video illustrating the experimental results accompanies the paper.

\section{Problem Setting and Design Rationale}
\label{sec:problem}

A hybrid force/motion controller requires, at each time instant, a task frame
and knowledge of the interaction force.
In the considered setting, these quantities are not assumed from an environment
model nor measured by an F/T sensor during the control task.
Instead, they are inferred from the image produced by a soft optical tactile
sensor.
This image encodes the deformation of the elastomer, whose relation to contact
geometry and force depends on the pad mechanics, the optical path, and the
morphology of the internal pins.
An accurate closed-form inverse of this mapping is not available and therefore
is extracted from data, and used in real-time control to estimate the observable
contact variables.

\begin{assumption}[Planar, featureless contact]
  \label{ass:plane}
  \looseness=-1 The contact surface is locally planar and featureless, its
  extent is large with respect to the contact patch of the tactile sensor.
  Its edges do not enter the sensing area.
\end{assumption}

Under~\cref{ass:plane}, the local task frame associated with the contact plane
is determined only up to transformations that leave the tactile image unchanged.
Different positions of the sensor over a featureless plane and different tilted
directions forming the same angle with the normal to the plane do not modify the
local geometry observed by the tactile sensor.

Thus, the identifiable parts of the contact are the signed offset $d$ along the
surface normal $\bm n$, together with the two angles that determine the
direction of $\bm n$.
Stated differently, among the relative sensor-surface pose variables, touch can
resolve the indentation depth and the two tilt angles between the surface normal
and the approach direction of the sensor.
The tangential position and the twist angle about the normal are structurally
unobservable from a single tactile contact with a locally planar, featureless
surface~\cite{2020-LepLlo,2021-LepLlo}.

\begin{assumption}[Negligible contact moment]
  \label{ass:moment}
  The moment transmitted at the contact interface is negligible over the
  intended operating range, because the contact patch is small relative to the
  characteristic scale of the task.
\end{assumption}

A general contact wrench has six components, three forces and three torques.
Under \cref{ass:moment}, however, the contact information retained by the
measurement model is restricted to the three force components.
No contact moment is therefore available for regulation of the torque
components.

The two assumptions above determine the force/motion decompositions considered
in this work.
Out of the set of task-space force/motion directions, only the observable ones
are subject to control, which discards three components from motion control,
namely the two tangential in-plane directions and the angular twist about the
normal to the plane.
Moreover, due to \cref{ass:moment}, the three components of the moment are
necessarily set to zero.
As a result, the
\looseness=-1 six remaining directions that can be controlled admit the
decompositions summarized in \cref{tab:ledger}.
Since the tactile measurement provides indentation depth, two surface-to-sensor
tilt angles, and a 3D contact force, the implemented controller assigns forces
in the three directions of the contact task frame and two degrees of angular
motion to align the sensor with the estimated surface normal.
The remaining twist angle about the normal is completed by task augmentation.

\begin{table}[!t]
  \centering
  \tabstyle
  \caption{Task-space decompositions induced by the local planar-contact model
    and realized in this work.
    The second column counts the number of directional components.
    The third column lists the nature of those that are observable from sensing
    under \cref{ass:plane}.
    The fourth column lists the quantities that are estimated under
    \cref{ass:moment}.
    The last column defines the chosen control task.
  }
  \label{tab:ledger}
  \begin{tabularx}{\linewidth}{@{}l c L{0.85}L{1.05}L{1.10}@{}}
    \toprule
    \textbf{Direction} & \textbf{\#} & \textbf{Observable}
    & \textbf{Estimated} & \textbf{Controlled} \\
    \midrule
    Tangential & $2$ & Force & Force & Force \\
    Normal     & $1$ & Force/motion & Force/motion & Force \\
    Tilt       & $2$ & Force/motion & Motion & Motion \\
    Twist      & $1$ & Force & -- & Task augmentation \\
    \bottomrule
  \end{tabularx}
\end{table}

The normal and tangential force directions are not equivalent from a contact
mechanics point of view.
Tangential forces can be transmitted only within the friction cone, whose size
is determined by the normal load.
In formulating the regulation problem, the admissible range of tangential force
depends on the normal force reference.
In the frictionless limit, a planar contact would support only a normal force,
while the tangential directions would be directions of free motion.
Therefore, the present formulation assumes that sufficient friction is present,
so that the tangential force components considered in \cref{tab:ledger} are
physically realizable within the operating range.

\looseness=-1 The notation used in this paper is fixed as follows.
The setting involves a fixed world frame $W$, a tool frame $T$ at the tactile
sensor tip, and a contact task frame $C$ whose $z$-axis is aligned with the
contact normal.
The F/T sensor frame is aligned with the tool frame $T$ (although displaced
along the $z$-direction), and bears thus no extra label.
Vectors are expressed in the frame in which they are defined.
A left superscript indicates that the vector is expressed instead in a different
frame.
All rotation matrices describe orientations with respect to the world frame,
i.e., $\bm R = {}^{W}\!\bm R$.
Unless a subscript is present, rotation matrices refer to the orientation of the
tool frame.
A hat denotes a quantity inferred from tactile data, either regressed from a
single image or estimated by the state estimator.

\section{Self-Labeling Dataset Acquisition}
\label{sec:dataset}

This section presents the acquisition protocol used to learn the tactile mapping
defined in \cref{sec:problem}.
The protocol is designed to excite the observable contact variables while
varying the unobservable ones as nuisance variables.
Robot proprioception provides the geometric labels, and the auxiliary F/T sensor
provides ground-truth force labels.

\subsection{Physical Setup}
\label{subsec:setup}

\looseness=-1 The manipulator both produces the contact and provides the
proprioceptive quantities used for labeling.
In this work, a DigiTac SOT sensor is mounted at the tool tip of a Universal
Robots UR10 manipulator, with an ATI Mini45 F/T sensor placed in series between
the two (\cref{tab:setup,fig:data_acquisition_setup}).
The F/T sensor is used only during dataset acquisition to construct the force
labels; its measurements are not inputs to the regressor or to the online
control loop.
The DigiTac is a soft optical tactile sensor composed of a deformable fingertip
mounted on a rigid case~\cite{2022-LepLinMonLlo}.
Its skin is a consumable part and was replaced during the experimental campaign;
for this, the dataset spans three different skins.

The contact surface is a rigid, flat and horizontal metal plate.
Within the prescribed operational domain, the plate satisfies \cref{ass:plane}.
The horizontal mounting is an acquisition convenience because it allows the
commanded tool pose to be expressed directly relative to the surface.
For any other known and fixed surface pose, the same protocol applies after a
suitable rotation.

\begin{table}[!t]
  \centering
  \tabstyle
  \caption{Experimental setup used for data acquisition and closed-loop
    validation.}
  \label{tab:setup}
  \begin{tabularx}{\linewidth}{@{}l
  l@{\hspace{4pt}}>{\raggedright\arraybackslash}X@{}}
    \toprule
    \textbf{Component} & \textbf{Model} & \textbf{Key specification} \\
    \midrule
    Manipulator & UR10 & {Repeatability} $0.1\,\mbox{mm}$ \\
    \midrule
    \multirow{2}{*}{F/T sensor} & \multirow{2}{*}{\shortstack[l]{ATI\\[1pt] Mini45}}
    & {Force resolution} $1/16\,\mbox{N} \approx \num{6.25e-2}\,\mbox{N}$ \\
    & & {Torque resolution} $1/752\,\mbox{Nm} \approx \num{1.33e-3}\,\mbox{Nm}$ \\
    \midrule
    \multirow{2}{*}{SOT sensor} & \multirow{2}{*}{DigiTac}
    & {Data acquisition} $640 \times 480$ @ $30\,\mbox{fps}$ \\
    & & {Real-time control} $320 \times 240$ @ $60\,\mbox{fps}$ \\
    \midrule
    \multirow{3}{*}{\makecell[l]{Tool\\assembly}} & \multirow{3}{*}{--}
    & {Mass} $m = 87\,\mbox{g}$ \\
    & & {Robot flange $\to$ SOT tip} $\bm r = (0,\,0,\,119.5)\,\mbox{mm}$ \\
    & & {F/T sensor $\to$ SOT tip} $\bm l = (0,\,0,\,85.0) \,\mbox{mm}$ \\
    \bottomrule
  \end{tabularx}
\end{table}

\input{assets/data-acquisition-setup}

\subsection{Sampling Design}
\label{subsec:sampling}

Each sample of the dataset pairs one tactile image with the quantities declared
observable in \cref{sec:problem}: the scalar indentation depth, the two
surface-to-sensor tilt angles, and the 3D contact force.
The sampling protocol spans these quantities while also varying the unobservable
tangential position and normal twist.
These variables are deliberately omitted from the labels but randomized to
reduce the risk that the regressor learns acquisition-specific correlations.

Consider a world frame whose $z$-axis is opposite to gravity and parallel to the
normal $\bm n$ of the horizontal surface.
Let the target tool position be $\tilde{\bm p}=(\tilde x,\tilde y,\tilde z)$ and
its orientation represented by the XYZ Roll-Pitch-Yaw angles $(\tilde\phi,
\tilde\theta, \tilde\psi)$ as
\begin{equation}
  \tilde{\bm R}
  =
  \bm R_z(\tilde\psi)\,
  \bm R_y(\tilde\theta)\,
  \bm R_x(\tilde\phi).
  \label{eq:tilt_param}
\end{equation}
\looseness=-1 Because the plate is horizontal, $(\tilde\phi,\tilde\theta)$
determine the surface-to-sensor tilt, up to the $\pi$ rotation offset associated
with the approach direction, and can therefore be directly used as geometric
labels.
One target pose sample is obtained as follows.
\begin{enumerate}[leftmargin=*]
  \item The position is drawn from an admissible box aligned with the surface,
  \begin{equation}
    \tilde{\bm p}\in
    \mathcal B
    =
    [-\bar x,\bar x]
    \times[-\bar y,\bar y]
    \times[\underline z,\bar z].
  \end{equation}
  The depth $\tilde z$ varies the indentation, with $\underline z<0$ and $\bar z
  \geq 0$, whereas the lateral displacement $(\tilde x,\tilde y)$ excites the
  tangential force channels while varying the unobservable contact location.

  \item The yaw angle is drawn from $\tilde\psi\in[-\bar\psi,\bar\psi]$, thereby
    varying the unobservable twist about the surface normal.

  \item The roll and pitch angles are drawn as $\tilde\phi\in[0,2\pi)$ and
    $\tilde\theta\in(-\pi/2,\pi/2]$.
  The resulting rotation is retained only if the tool $z$-axis lies inside an
  admissible cone with opening angle $\bar \gamma$ around the surface normal:
  \begin{equation}
    \bm e_3^\top\tilde{\bm R}\,\bm e_3
    =
    \cos \tilde\phi \cos \tilde\theta
    \leq
    -\cos \bar\gamma(\tilde z),
    \label{eq:approach_cone}
  \end{equation}
  where $\bm e_3 = (0,0,1)$;
  otherwise, the angles are resampled.

  \item A second roll-pitch pair $(\tilde\phi?{{pre}},\tilde\theta?{{pre}})$ is
    drawn under the same constraint.
  This pair determines the orientation from which the surface is approached and
  varies the loading path of the contact independently of the labeled final
  pose.
\end{enumerate}

For some combinations of tilt and indentation, the contact patch moves from the
sensing area toward the rim of the SOT dome.
To exclude these contacts from the operational domain, the admissible cone
opening angle decreases with depth from $\gamma_{\max}\in[0,\pi/2)$ according to
\begin{equation}
  \cos \bar\gamma(z)
  =
  \min\{0,z\}
  \,\frac{1-\cos \gamma_{\max} }{\underline z}
  +
  \cos \gamma_{\max}.
  \label{eq:cone_constraint}
\end{equation}

Each sample is acquired through a complete approach-release cycle, with the
target pose and pre-contact orientation sampled independently for each cycle.
Unloading the pad between samples reduces sequential dependence due to
persistent deformation, without requiring the stronger assumption that
consecutive samples are statistically uncorrelated.

\subsection{Acquisition Cycle}
\label{subsec:cycle}

Given a target $(\tilde{\bm p},\tilde{\bm R})$, define the pre-contact
orientation as
\begin{equation}
  \tilde{\bm R}?{{pre}}
  =
  \bm R_z(\pi)\,
  \bm R_y(\tilde\theta?{{pre}})\,
  \bm R_x(\tilde\phi?{{pre}}),
\end{equation}
where the rotation by $\pi$ around the fixed $z$-axis is introduced to align
(arbitrarily) the $y$-axis of the tool frame with the $y$-axis of the contact
frame.
The target is realized through the six phases shown in \cref{fig:acq_phases}.

\input{assets/acq-phases}

\begin{enumerate}[leftmargin=*]
  \item \textit{Home.} The tool moves above the surface with $h?{safe} > 0$ to
    $(h?{{safe}}\bm e_3,\bm R_z(\pi) \bm I_3 \bm R_x(\pi))$, allowing the
    tactile pad to relax before the next sample.

  \item \textit{Pre-rotate.} Holding the clearance position, the tool reorients
    to $(h?{{safe}}\bm e_3,\tilde{\bm R}?{{pre}})$.
  Once the robot and sensor readings have settled, the out-of-contact F/T
  measurement $(\bm f?{{pre}},\bm\mu?{{pre}})$ is recorded together with the
  attained roll and pitch $(\phi?{{pre}},\theta?{{pre}})$.
  This measurement is used offline to compensate for gravity and for the slowly
  varying F/T bias.

  \item \textit{Indent.} The tool approaches along the surface normal to
    $(\tilde z\bm e_3,\tilde{\bm R}?{{pre}})$, establishing contact without a
    commanded in-plane displacement.

  \item \textit{Shear.} While maintaining contact, the tool translates in-plane
    to $(\tilde{\bm p},\tilde{\bm R}?{{pre}})$.
  This displacement excites the tangential force channels.

  \item \textit{Post-rotate.} Still in contact, the tool rotates to $(\tilde{\bm
    p},\tilde{\bm R})$.
  Reaching the target tilt after indentation varies the loading path
  independently of the final labeled pose.
  Once the readings have settled, tactile image, attained robot pose, and F/T
  measurement are recorded together.

  \item \textit{Raise.} The tool retracts along the surface normal to $((\tilde
    x,\tilde y,h?{{safe}}),\tilde{\bm R})$, releasing contact without a
    commanded in-plane dragging motion.
\end{enumerate}

Phases 3--5 are executed at the speed which is $k?{{safe}} < 1$ times the
maximum speed allowed by the UR10 to limit impact and inertial effects.
The clearance height $h?{{safe}} > 0$ separates the {\em Home} and {\em Raise}
poses from the plate.
The numerical values used for each dataset batch are reported in
\cref{tab:exp_params}.

\begin{table}[!t]
  \centering
  \tabstyle
  \caption{Parameters for the data acquisition routine.}
  \label{tab:exp_params}
  \begin{tabularx}{\linewidth}{@{}p{1.6cm} X C{0.4} C{0.4}@{}}
    \toprule
    \textbf{Scope} & \textbf{Description} & \textbf{Symbol} & \textbf{Value} \\
    \midrule
    \multirow{4}{*}{\makecell[l]{All batches \\ ($10 \, 000$)}}
    & Safe-speed factor & $k?{safe}$ & $0.2$ \\
    & Post-contact yaw bound & $\bar{\psi}$ & $15\,\mbox{deg}$ \\
    & Safe clearance height & $h?{safe}$ & $20\,\mbox{mm}$ \\
    & Lateral translation bounds & $\bar{x}, \bar{y}$ & $5\,\mbox{mm}$\\
    \midrule
    \multirow{3}{*}{\makecell[l]{Batch $1$ \\ ($5 \, 750$)}}
    & Max contact-cone opening & $\gamma?{max}$ & $10\,\mbox{deg}$ \\
    & Indent depth lower bound & $\underline{z}$ & $-2\,\mbox{mm}$ \\
    & Indent depth upper bound & $\bar{z}$ & $0\,\mbox{mm}$ \\
    \midrule
    \multirow{3}{*}{\makecell[l]{Batch $2$ \\ ($2 \, 500$)}}
    & Max contact-cone opening & ${\gamma}?{max}$ & $25\,\mbox{deg}$ \\
    & Indent depth lower bound & $\underline{z}$ & $-2.2\,\mbox{mm}$ \\
    & Indent depth upper bound & $\bar{z}$ & $0.5\,\mbox{mm}$ \\
    \midrule
    \multirow{3}{*}{\makecell[l]{Batch $3$ \\ ($1 \, 750$)}}
    & Max contact-cone opening & ${\gamma}?{max}$ & $5\,\mbox{deg}$ \\
    & Indent depth lower bound & $\underline{z}$ & $-3\,\mbox{mm}$ \\
    & Indent depth upper bound & $\bar{z}$ & $0.5\,\mbox{mm}$ \\
    \bottomrule
  \end{tabularx}
\end{table}

\datasetstatement

\subsection{Label Extraction}
\label{subsec:labels}

\looseness=-1 The geometric labels are obtained from the attained robot pose
relative to the known plate frame.
The force label is instead computed from the F/T measurements.
The raw F/T reading includes the contact wrench, the weight of the downstream
tool assembly, and a slowly varying bias, all expressed in the tool frame.
These contributions are separated using the out-of-contact measurement recorded
during the \textit{Pre-rotate} phase.

Let $m$ be the mass of the downstream assembly, including the SOT sensor and its
mount, and $g$ the gravity acceleration.
At the pre-contact orientation $\bm R?{{pre}}$, the F/T sensor reads
\begin{equation}
  \label{eq:f_pre}
  \bm f?{{pre}}
  =
  -mg\bm R?{{pre}}^\top\bm e_3+\bm b_f,
\end{equation}
where $\bm f?{{pre}}$ and the bias $\bm b_f$ are expressed in the tool frame.

At the labeled contact configuration, the measured force is
\begin{equation}
  \label{eq:f_post}
  \bm f
  =
  \bm f?{{gt}}
  -mg\bm R^\top\bm e_3+\bm b_f,
\end{equation}
where $\bm f?{{gt}}$ is the contact force, also expressed in the tool frame.
Assuming the bias approximately constant over one acquisition cycle, subtraction
of \cref{eq:f_pre} from \cref{eq:f_post} cancels it, yielding
\begin{equation}
  \bm f?{{gt}}
  =
  \bm f-\bm f?{{pre}}
  +mg(\bm R-\bm R?{{pre}})^\top\bm e_3,
  \label{eq:preproc_force}
\end{equation}
which is used in the final label of the sample.

The tactile images are converted to grayscale and downsampled before training,
reducing the computational cost of training and inference.
The same image format is used during training and evaluation, and later for
control.

\section{Tactile Measurement Regressor}
\label{sec:network}

\looseness=-1 A tactile image represents a spatially structured deformation
field, in which information is encoded by the displacement of the pins.
Convolution provides a natural inductive bias for this type of signal and
therefore forms the basis of the proposed regressor.

\subsection{Architecture}

The proposed model (\cref{fig:cnn_structure}) is a ResNet-style
regressor~\cite{2016-HeZhaRenSun}.
It combines a residual convolutional backbone, which learns a representation of
the contact, with two parallel multilayer perceptron (MLP) heads that predict
the mean and log-variance of the tactile measurement in $\mathbb{R}^{6}$,
respectively.

\input{assets/cnn-structure}

\looseness=-1 The architecture combines elements whose advantages are
established in the literature.
Residual connections are known to ease the optimization of deep convolutional
backbones~\cite{2016-HeZhaRenSun}.
Starting from a residual backbone with four stages, candidate modifications were
introduced one at a time and retained when they reduced the validation loss.
Examples include group normalization~(GN)~\cite{2018-WuHe} in place of batch
normalization, squeeze-and-excitation~(SE)~\cite{2018-HuSheSun}, trunk depth and
width, size of the pooled feature map, width of the hidden layers, dropout rate,
and optimization hyperparameters such as the learning rates.
Since each retained change became the baseline for the next, these comparisons
are not independent and are therefore not reported as an ablation study.
The final architecture is described in detail below.

\paragraph{Backbone}
The backbone is deliberately kept small to support real-time deployment.
The input tactile image is normalized to the range $[-0.5,0.5]$ and passed
through a shallow stem comprising a $3\times3$ convolution with stride~$2$, GN,
and a rectified linear unit (ReLU).
This operation halves the spatial resolution to reduce the computational cost,
while retaining a sufficiently small kernel to preserve local deformation
patterns.
Four residual stages follow.
Each stage comprises two blocks, except for the last stage, which comprises a
single block.
The stage widths are $32$, $64$, $128$, and $128$ channels, and the spatial
resolution is reduced by $2\times2$ max-pooling.
Adaptive average pooling produces a $128\times8\times6$ feature map at the
output of the backbone.
The backbone accounts for $1.0\times10^{6}$ of the $4.15\times10^{6}$ parameters
of the complete model.

\paragraph{Residual Blocks}
Each residual block implements the mapping
\begin{equation}
  \bm{Y}
  =
  \operatorname{ReLU}\!\left(
  \mathcal{F}(\bm{X}) \cdot \alpha(\bm{X})
  + \mathcal{S}(\bm{X})
  \right),
\end{equation}
where $\bm X$ and $\bm Y$ are the input and output feature maps of the block,
and $\cdot$ denotes channel-wise scaling.
The residual function $\mathcal{F}(\cdot)$ comprises two $3\times3$ convolutions
with GN using eight groups and an intermediate ReLU.
The skip path $\mathcal{S}(\bm{X})$ is either an identity mapping or a
$1\times1$ convolution when channel expansion is required.
A lightweight SE block~\cite{2018-HuSheSun}, denoted by $\alpha(\cdot)$ and
using a reduction ratio of~$8$, recalibrates the features channel-wise before
the residual addition.

\paragraph{Regression Heads}
The pooled features are flattened into a shared $6144$-dimensional vector.
After $20\%$ dropout, this vector is processed by two parallel MLP heads, each
with dimensions $6144 \rightarrow 256 \rightarrow 6$ and with ReLU and $20\%$
dropout between the two linear layers.
The two heads produce the mean prediction and the per-sample log-variance of the
tactile measurement, respectively.

\paragraph{Output}
The mean head predicts the tactile measurement
\begin{equation}
  \label{eq:output_vector}
  \hat{\bm{y}}
  =
  \begin{pmatrix}
    \hat{z} &
    \hat{\phi} &
    \hat{\theta} &
    ({}^T\hat{\!\bm{f}}?{c})^\top
  \end{pmatrix}^{\top}
  \in \mathbb{R}^{6},
\end{equation}
\looseness=-1 where ${}^T\hat{\!\bm{f}}?{c}\in\mathbb{R}^{3}$ is the regressed
contact force expressed in the tool frame to minimize cross-talk among the
regression channels.

The variance head predicts a per-sample log-variance vector
$\hat{\bm{s}}\in\mathbb{R}^{6}$, where $\hat{s}_{i}=\log(\hat{\sigma}_{i}^{2})$.
During inference, $\hat{\bm{s}}$ is exponentiated element-wise to construct the
per-sample covariance
\begin{equation}
  \label{eq:covariance}
  \hat{\bm{W}}
  =
  \diag\!\left(\exp(\hat{\bm{s}})\right).
\end{equation}
\looseness=-1 The contact state estimator of \cref{sec:control} requires a
measurement covariance at each update, and the heteroscedastic head provides
this covariance on a per-sample basis.
The joint prediction of an estimate and its confidence, followed by their
propagation through a Bayesian filter, follows the scheme established
in~\cite{2021-VazLep}.

\looseness=-1 The reduction from the general
$\mathrm{SE}(3)\times\mathbb{R}^{6}$ space is motivated by the decompositions
derived in \cref{sec:problem}.
On a locally planar surface, the tangential position and the twist angle around
the surface normal are unobservable under~\cref{ass:plane}, and the contact
moment is negligible under~\cref{ass:moment}.

\subsection{Training Methodology}
\label{subsec:training}

The network is trained on the collected dataset of $10\,000$ labeled samples,
partitioned at random into training, validation, and test subsets in the
proportions $70\%/15\%/15\%$ reported in \cref{tab:hyperparams}.
It keeps model selection, performed on the validation subset, separate from the
final evaluation, performed on the test subset once the training configuration
is fixed.
The training procedure is described below.

\paragraph{Loss Function}
The regression problem is formulated as maximum-likelihood estimation under a
diagonal Gaussian observation model.
For each output dimension~$i$, the predicted distribution is
$\mathcal{N}(\hat{y}_{i},\exp(\hat{s}_{i}))$, where $\hat{s}_{i}$ is the
predicted log-variance and $y_i$ denotes the corresponding label.
Following~\cite{2017-KenGal}, the resulting Gaussian negative log-likelihood
(NLL) loss is
\begin{equation}
  \mathcal{L}
  =
  \frac{1}{6}
  \sum_{i=1}^{6}
  \left(
  \frac{1}{2}
  \exp(-\hat{s}_{i})
  (\hat{y}_{i}-y_{i})^{2}
  +
  \frac{1}{2}\hat{s}_{i}
  \right).
  \label{eq:nll_loss}
\end{equation}

\begin{table}[!t]
  \centering
  \tabstyle
  \caption{Training configuration and hyperparameters.}
  \label{tab:hyperparams}
  \begin{tabularx}{\linewidth}{@{}l@{\hspace{6pt}}L{1.3}@{\hspace{-9pt}}C{0.7}@{\hspace{-5pt}}}
    \toprule
    \textbf{Scope} & \textbf{Description} & \textbf{Value} \\
    \midrule
    \multirow{2}{*}{Dataset}
    & Split (train/validation/test) & $70\%/15\%/15\%$ \\
    & Image dimensions & $320\times240$ \\
    \midrule
    \multirow{4}{*}{Data augmentation}
    & Application probability & $50\%$ \\
    & Contrast-factor interval & $\pm10\%$ \\
    & Brightness-factor interval & $\pm5\%$ \\
    & Gaussian-noise standard deviation & $0.02$ \\
    \midrule
    \multirow{6}{*}{Optimization}
    & Trainable parameters & \num{4.15e6} \\
    & Optimizer & AdamW \\
    & Weight decay & \num{5.0e-3} \\
    & Batch size & $256$ \\
    & Base learning rate (Stage~1) & \num{1.0e-3} \\
    & Base learning rate (Stage~2) & \num{1.0e-4} \\
    \midrule
    \multirow{6}{*}{Scheduling}
    & Warmup strategy & Linear (10 epochs) \\
    & Decay strategy & Reduction on plateau \\
    & Decay factor & $0.5$ \\
    & Decay patience & 5 epochs \\
    & Maximum number of epochs & $500$ \\
    & Early-stopping patience & 50 epochs \\
    \bottomrule
  \end{tabularx}
\end{table}

\paragraph{Two-Stage Training}
\looseness=-1 Training proceeds in two stages.
During \emph{homoscedastic pretraining}, the variance head is frozen, while a
single global log-variance vector shared across all input samples is optimized
jointly with the remainder of the network as an auxiliary learnable parameter.
Warming up the mean while the variance is held fixed has been shown to prevent
the convergence failures instead observed when both outputs are optimized from
initialization~\cite{2024-SluCatHes}.
The failures arise because the NLL loss suppresses the gradient on poorly fitted
samples, yielding a stable yet inaccurate mean
predictor~\cite{2022-SeiTavAntMar}.

\emph{Heteroscedastic fine-tuning} then starts from the Stage~1 early-stopping
checkpoint and optimizes all parameters end-to-end, including the log-variance,
using a reduced learning rate.
In this stage, the network learns input-dependent, per-sample uncertainty
estimates, assigning a larger predicted log-variance in $\hat{\bm{s}}$ to
ambiguous or unusual tactile images.

Both single-stage alternatives were evaluated during development.
Homoscedastic training alone converges reliably but can only return a single
global variance, missing the per-sample uncertainty requirement.
On the other hand, heteroscedastic training did not converge reliably because of
the aforementioned issues.
Both stages are therefore necessary.
Stage~1 learns the global log-variance rather than fixing it as suggested
in~\cite{2024-SluCatHes}, so that the loss scale of each output channel is
calibrated before per-sample variances are enabled.

\paragraph{Optimization}
Online data augmentation combines photometric distortions, representing
variations in illumination, with additive Gaussian noise, representing
electronic noise.
The model is optimized using AdamW~\cite{2019-LosHut}.
The learning rate increases linearly from zero during the first epochs and is
subsequently reduced when the validation loss reaches a plateau.
Training stops when the validation loss does not improve for $50$ consecutive
epochs, and the parameters attaining the lowest validation loss are retained.
\Cref{tab:hyperparams} reports the complete configuration.

\section{Tactile-Driven Hybrid Force/Motion Control}
\label{sec:control}

The control objective is to regulate a prescribed contact force against a
surface whose pose is unknown and slowly time-varying, while keeping the
alignment of the tactile sensor with the local surface normal.
Touch is the only source of exteroceptive feedback available to the controller.

\begin{figure*}[!t]
  \centering
  \resizebox{\textwidth}{!}{%
    \begin{tikzpicture}[
      x=1cm,
      y=1cm,
      block/.style={
        figblock,
        sharp corners,
        minimum width=2.75cm,
        minimum height=1.55cm,
        text width=2.55cm,
        align=center,
        inner xsep=4pt,
        inner ysep=5pt
      },
      wideblock/.style={
        block,
        minimum width=3.15cm,
        text width=2.95cm
      },
      every node/.style={font=\small},
      commandblock/.style={fighue=nordOrange},
      estimationblock/.style={fighue=nordPurple},
      physicalblock/.style={fighue=nordGreen},
      controlline/.style={
        figline=nordOrange,
        densely dashed,
        line width=1pt
      },
      estimationline/.style={
        figline=nordPurple,
        line width=1pt
      },
      interactionline/.style={
        figline=nordGreen,
        line width=1.1pt
      },
      signallabel/.style={
        draw=none,
        fill=none,
        inner xsep=1pt,
        inner ysep=1pt,
        font=\footnotesize
      },
      junction/.style={
        circle,
        fill=nordPurple!75!black,
        inner sep=1.5pt
      }
      ]

      \node[wideblock, commandblock] (target) at (-0.2,3.3) {
        \textbf{Target Generator}\\[2pt]
        \scriptsize Hybrid force and orientation target\\[2pt]
        \scriptsize $50\,\mbox{Hz}$
      };

      \node[wideblock, commandblock] (orientation) at (4.15,4.25) {
        \textbf{Orientation Filter}\\[2pt]
        \scriptsize Smooth alignment reference\\[2pt]
        \scriptsize \cref{eq:ref_orientation,eq:att_admittance}
        \quad $50\,\mbox{Hz}$
      };

      \node[wideblock, commandblock] (position) at (4.15,2.35) {
        \textbf{Position Admittance}\\[2pt]
        \scriptsize Force-to-motion conversion\\[2pt]
        \scriptsize \cref{eq:pos_admittance}
        \quad $50\,\mbox{Hz}$
      };

      \node[wideblock, commandblock] (motion) at (8.55,3.3) {
        \textbf{Motion Controller}\\[2pt]
        \scriptsize Cartesian pose tracking\\[2pt]
        \scriptsize \cref{eq:motion_control}
        \quad $20\,\mbox{Hz}$
      };

      \node[wideblock, physicalblock] (robot) at (12.55,3.3) {
        \textbf{Manipulator}\\[2pt]
        \scriptsize Robot and tactile tool
      };

      \node[wideblock, physicalblock] (environment) at (17.0,3.3) {
        \textbf{Environment}\\[2pt]
        \scriptsize Unknown, slowly\\
        \scriptsize varying surface
      };

      \node[wideblock, estimationblock] (ekf) at (4.15,-0.85) {
        \textbf{Contact State EKF}\\[2pt]
        \scriptsize Tactile-proprioceptive fusion\\[2pt]
        \scriptsize
        \cref{eq:ekf_process,eq:h_function}
        \quad $50\,\mbox{Hz}$
      };

      \node[wideblock, estimationblock] (detector) at (8.55,-0.85) {
        \textbf{Contact Detection}\\[2pt]
        \scriptsize Schmitt trigger on $\hat z$\\[2pt]
        \scriptsize $50\,\mbox{Hz}$
      };

      \node[wideblock, estimationblock] (regressor) at (12.55,-0.85) {
        \textbf{Tactile Regressor}\\[2pt]
        \scriptsize Instantaneous measurement\\[2pt]
        \scriptsize \cref{sec:network}
        \quad $50\,\mbox{Hz}$
      };

      \node[
      figgroup,
      dotted,
      very thick,
      inner xsep=8pt,
      inner ysep=8pt,
      fit=(orientation)(position)
      ] (admittancegroup) {};

      \node[
      font=\bfseries\small,
      fill=white,
      inner xsep=4pt,
      anchor=south east
      ] at ([xshift=0.35cm,yshift=1pt]admittancegroup.north east) {Admittance
      Layer};

      \coordinate (measurementlaneR) at ([yshift=-0.6cm]regressor.south);
      \coordinate (measurementlaneL) at ([xshift=0.0cm,yshift=-0.6cm]ekf.south);

      \node[
      figgroup,
      dotted,
      very thick,
      inner xsep=8pt,
      inner ysep=8pt,
      fit=(ekf)(detector)(regressor)(measurementlaneR)(measurementlaneL)
      ] (estimationgroup) {};

      \node[
      font=\bfseries\small,
      fill=white,
      inner xsep=4pt,
      above=1pt of estimationgroup.north
      ] {Tactile Contact State Estimation};

      \draw[controlline]
      ([yshift=0.35cm]target.east)
      --
      node[
      signallabel,
      pos=1,
      anchor=south east,
      xshift=4pt,
      yshift=20pt
      ]{
        $\phi^\star,\,\theta^\star,\,\psi^\star$
      }
      ++(0.60,0)
      |- (orientation.west);

      \draw[controlline]
      ([yshift=-0.35cm]target.east)
      --
      node[
      signallabel,
      below=20pt,
      xshift=6pt
      ]{
        $\bm f?{c}^{\star}$
      }
      ++(0.60,0)
      |- (position.west);

      \draw[controlline]
      (orientation.east)
      -- ++(0.50,0)
      |-
      node[
      signallabel,
      pos=0.72,
      above=20pt
      ]{
        $\tilde{\bm R},\,\tilde{\bm\omega}$
      }
      ([yshift=0.35cm]motion.west);

      \draw[controlline]
      (position.east)
      -- ++(0.50,0)
      |-
      node[
      signallabel,
      pos=0.72,
      below=20pt
      ]{
        $\tilde{\bm p},\,\tilde{\bm v}$
      }
      ([yshift=-0.35cm]motion.west);

      \draw[controlline]
      (motion.east)
      --
      node[
      signallabel,
      above=3pt
      ]{
        $\bm v,\,\bm\omega$
      }
      (robot.west);

      \draw[interactionline, <->]
      (robot.east)
      --
      node[
      signallabel,
      above=3pt
      ]{
        contact
      }
      (environment.west);

      \draw[estimationline]
      ([xshift=0.70cm]robot.south)
      --
      node[
      signallabel,
      right=3pt
      ]{
        $\mathcal I$
      }
      ([xshift=0.70cm]regressor.north);

      \draw[estimationline]
      (regressor.west)
      --
      node[
      signallabel,
      above=3pt
      ]{
        $\hat z$
      }
      (detector.east);

      \draw[estimationline]
      (detector.west)
      --
      node[
      signallabel,
      above=3pt
      ]{
        $c$
      }
      (ekf.east);

      \draw[estimationline]
      (regressor.south)
      -- (measurementlaneR)
      --
      node[
      signallabel,
      above=3pt
      ]{
        $\hat{\bm y},\,\hat{\bm W}$
      }
      (measurementlaneL)
      -- ([xshift=0.0cm]ekf.south);

      \coordinate (midlevel) at
      ($([xshift=-0.70cm]ekf.north)!0.5!([xshift=-0.70cm]position.south)$);

      \coordinate (poseorigin) at
      ([xshift=-0.70cm]robot.south);

      \coordinate (posebranch) at
      (poseorigin |- midlevel);

      \coordinate (posemotionbranch) at
      (motion.south |- posebranch);

      \coordinate (poseekfbranch) at
      ([xshift=0.70cm]ekf.north |- posebranch);

      \draw[estimationline]
      (poseorigin)
      -- (posebranch)
      -- (poseekfbranch)
      --
      node[
      signallabel,
      pos=0.33,
      left=3pt
      ]{
        $\bm p,\,\bm R$
      }
      ([xshift=0.70cm]ekf.north);

      \draw[estimationline]
      (posemotionbranch)
      --
      node[
      signallabel,
      right=3pt
      ]{
        $\bm p,\,\bm R$
      }
      (motion.south);

      \node[junction] at (posemotionbranch) {};

      \draw[estimationline]
      ([xshift=-0.70cm]ekf.north)
      --
      node[
      signallabel,
      pos=0.66,
      right=3pt
      ]{
        $\hat{\bm R}_C, \, \hat{\!\bm f}?{c}$
      }
      ([xshift=-0.70cm]position.south);

      \coordinate (frameport) at ([xshift=-0.70cm]orientation.north);

      \coordinate (frametop) at ([yshift=1.0cm]orientation.north);

      \coordinate (framelane) at
      ([xshift=-0.40cm]target.west |- midlevel);

      \draw[estimationline]
      (midlevel)
      -- (framelane)
      -- (framelane |- frametop)
      --
      node[
      signallabel,
      pos=0.935,
      below=3pt
      ]{
        $\hat{\bm R}_C$
      }
      (frameport |- frametop)
      -- (frameport);

      \node[junction] at (midlevel) {};

      \node[
      figlegend,
      anchor=south,
      font=\footnotesize,
      align=center
      ] at (13.15,5.05) {%
        \tikz[baseline=-0.6ex]
        \draw[controlline] (0,0)--(0.45,0);\,reference/command%
        \quad
        \tikz[baseline=-0.6ex]
        \draw[estimationline] (0,0)--(0.45,0);\,sensing/estimation%
        \quad
        \tikz[baseline=-0.6ex]
        \draw[interactionline, <->] (0,0)--(0.6,0);\,physical interaction%
      };

    \end{tikzpicture}%
  }

  \caption{Tactile-driven hybrid force/motion control architecture.
    The tactile regressor maps each image $\mathcal I$ to the instantaneous
    measurement $\hat{\bm y}$ and covariance $\hat{\bm W}$.
    A Schmitt trigger applied to $\hat z$ provides the contact flag $c$ that
    enables the EKF measurement update.
    The EKF combines valid tactile measurements with robot proprioceptive pose
    data $(\bm p,\bm R)$ of the tool to continuously update the estimates of
    contact task frame $\hat{\bm R}_C$ and interaction force $\hat{\!\bm
    f}?{c}$.
    The robot pose is also supplied to the motion controller.
    The state $\bm \xi$ of the admittance model in \cref{eq:pos_admittance} is
    inside the position admittance block.
    The estimated contact variables close the loops of force regulation and
    surface alignment.
  }
  \label{fig:exp_control_architecture}
\end{figure*}
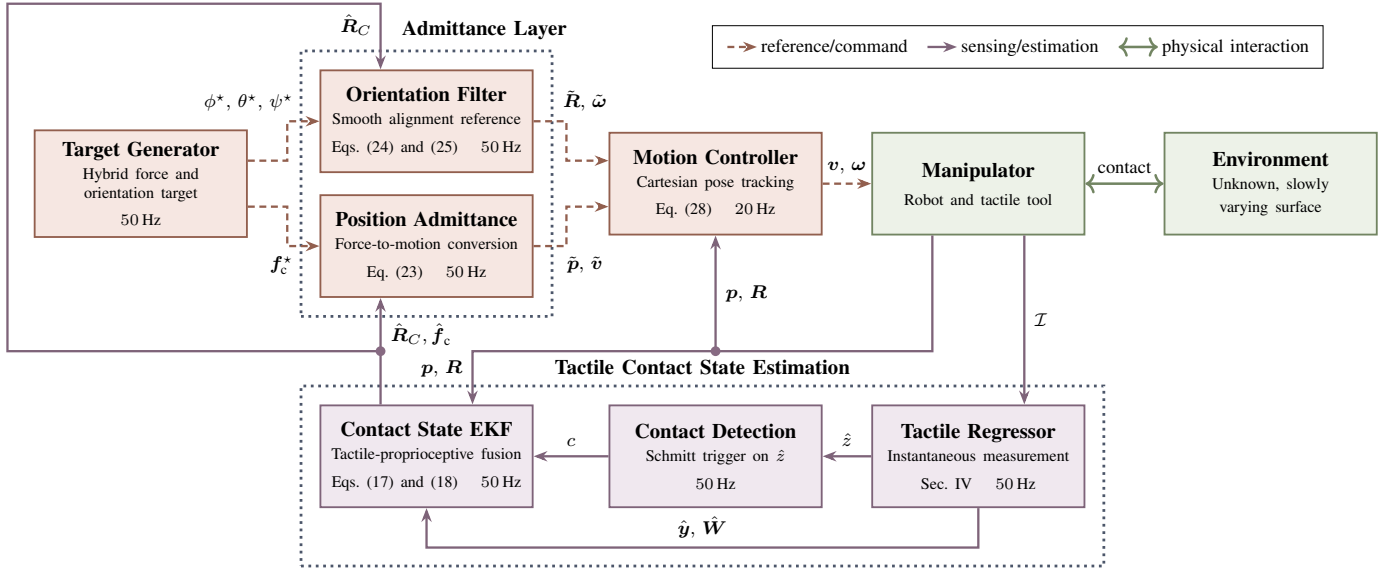

The system consists of a fixed-base manipulator equipped with the SOT sensor at
its tool tip, using the same mechanical configuration adopted for data
acquisition (\cref{fig:data_acquisition_setup}).
The F/T sensor is still present but no longer used for control.
In the architecture shown in \cref{fig:exp_control_architecture},
an Extended Kalman Filter (EKF) \cite{1982-AndMooEsl} fuses each instantaneous
tactile measurement with robot proprioceptive data to maintain a continuously
updated estimate of contact task frame and interaction force.
A target generator issues the hybrid target, composed by the desired contact
force and the desired sensor orientation.
An admittance layer combines these targets with the estimated contact task frame
and converts them into smooth motion references, which are tracked by a
geometric Cartesian velocity controller.

\subsection{Estimation of the Contact State}

\looseness=-1 The regressor of \cref{sec:network} returns only the instantaneous
tactile measurement $\hat{\bm y}$ expressed relative to the sensor.
Hybrid force/motion control instead requires a continuously updated
representation of the surface geometry and interaction force in a suitable frame
for decoupled control~\cite{2025-SicVilOriDeL}.
The method obtains this representation by fusing the tactile measurement with
the tool pose $(\bm p,\bm R)$ expressed in the world frame, as provided by the
robot interface.

The EKF updates the contact state estimate
\begin{equation}
  \label{eq:state_vector}
  \hat{\bm x}
  =
  \begin{pmatrix}
    \hat d &
    \hat\alpha &
    \hat\beta &
    \hat{\!\bm f}?{c}^{\top}
  \end{pmatrix}^{\top}
  \in\mathbb{R}^{6},
\end{equation}
where $\hat d$ is the signed offset of the estimated contact plane,
$(\hat\alpha,\hat\beta)$ parameterize its normal $\hat{\bm n}$ expressed in the
world frame, and $\,\hat{\!\bm f}?{c}\in\mathbb{R}^{3}$ is the related contact
force expressed in the estimated contact task frame.
Specifically,
\begin{equation}
  \label{eq:surface_normal}
  \hat{\bm n}
  =
  \begin{pmatrix}
    \cos \hat\alpha \sin \hat\beta &
    \sin \hat\alpha \sin \hat\beta &
    \cos \hat\beta
  \end{pmatrix}^{\top}
\end{equation}
and the corresponding plane is described by
\begin{equation}
  \hat{\bm n}^{\top}\bm q-\hat d=0,
  \label{eq:estimated_plane}
\end{equation}
for any point $\bm q$ on the plane.
This convention makes the first component of the tactile measurement model equal
to the signed sensor-to-plane displacement $\hat{\bm n}^{\top}\bm p-\hat d$.

The state in \cref{eq:state_vector} retains the part of the contact geometry
observable under the locally planar model of \cref{ass:plane}, namely the plane
offset and its normal, together with the three force components retained under
\cref{ass:moment}.
Thus, it does not include tangential contact coordinates, rotation about the
surface normal, and contact moment.

The orientation $\hat{\bm R}_C$ of the estimated contact task frame is
parameterized in moving axes as
\begin{equation}
  \label{eq:RC_param}
  \hat{\bm R}_C
  =
  \bm R_z(\hat\alpha)
  \bm R_y(\hat\beta)
  \bm R_z(\pi/2).
\end{equation}
By construction, the $z$-axis of $\hat{\bm R}_C$ coincides with $\hat{\bm n}$,
and the final rotation in \cref{eq:RC_param} makes its $x$-axis horizontal.

The parameterization in \cref{eq:surface_normal} is singular when $\sin
\hat\beta =0$, that is, when the estimated normal is aligned with the world
vertical axis.
Since we have considered in the closed-loop control experiments that the contact
surface is moving and rotating around a vertical lateral wall, this choice is
appropriate as $\hat\beta$ remains away from $0$ and $\pi$.
In any event, this singularity concerns only the selected local coordinates.

In the absence of a known model for the motion of the contact surface, the state
${\hat{\bm x}}$ in~\cref{eq:state_vector} evolves according to a random walk
process
\begin{equation}
  \label{eq:ekf_process}
  \begin{aligned}
    \dot{\hat{\bm x}}
    &=
    \bm Q^{1/2}\bm m_1\\
    \hat{\bm y}
    &=
    \bm h(\hat{\bm x},\bm p,\bm R)
    +
    \hat{\bm W}^{1/2}\bm m_2,
  \end{aligned}
\end{equation}
where $\bm m_i\sim\mathcal{N}(\bm 0,\bm I_6)$, $i = 1,2$, $\bm
Q\in\mathbb{R}^{6\times6}$ is a symmetric, positive semidefinite matrix, and
$\hat{\bm W}$ is the diagonal, positive definite matrix in~\cref{eq:covariance}.
Since this covariance matrix is provided for each tactile image by the variance
head described in~\cref{sec:network}, the EKF weights each tactile measurement
according to its associated covariance.
The tool pose $(\bm p,\bm R)$ in the measurement function $\bm h$ is treated as
a known, time-varying input to the measurement model provided by robot
proprioceptive information.

Expressing the estimated normal in the tool frame with \( {}^T \!
\hat{\bm{n}} = \bm R^{\top}\hat{\bm n} \), the measurement prediction evaluated
at the current state is
\begin{equation}
  \label{eq:h_function}
  \bm h(\hat{\bm x},\bm p,\bm R)
  =
  \begin{pmatrix}
    \hat{\bm n}^{\top}\bm p-\hat d
    \\[2mm]
    \pi+
    \arctan\!\left({}^T \! \hat n_y/{}^T \! \hat n_z\right)
    \\[2mm]
    -\arctan\!\left(
    {}^T \! \hat n_x/
    \sqrt{{}^T \! \hat n_y^2+{}^T \! \hat n_z^2}
    \right)
    \\[3mm]
    \bm R^{\top}
    \hat{\bm R}_C \,
    \hat{\!\bm f}?{c}
  \end{pmatrix}.
\end{equation}
The first component predicts indentation ${\hat z}$, the second and third
predict the surface-to-sensor tilt angles $(\hat \phi, \hat \theta)$, and the
last three express the estimated contact force ${}^T\hat{\!\bm{f}}?{c}$ in the
tool frame, all to be compared with the corresponding components of the
measurement $\hat{\bm y}$ in~\cref{eq:output_vector}.
Thus, \cref{eq:h_function} maps the persistent contact state, expressed in the
world frame, to the quantities relative to the sensor predicted by the tactile
regressor.

Contact is detected using a Schmitt trigger applied to the regressed indentation
$\hat z$.
Let two thresholds be defined as
\begin{equation}
  z*{{up}}?{{thr}}
  >
  z*{{lo}}?{{thr}}
  >
  0.
\end{equation}
Since penetration corresponds to negative indentation, contact is confirmed when
$\hat z<-z*{{up}}?{{thr}}$ and released when $\hat z>-z*{{lo}}?{{thr}}$.
The resulting hysteresis suppresses repeated switching in the proximity of
contact onset and detachment situations.

\looseness=-1 When contact is not detected, the EKF measurement update is
disabled because the regressor output does not represent a valid contact
measurement.
The state prediction is retained and the Kalman filter covariance matrix $\bm P$
is propagated according to the process model~\eqref{eq:ekf_process}.
Therefore, the latest estimated surface geometry remains available during
temporary losses of contact, while its uncertainty increases through the process
noise.

\subsection{Control Pipeline}

The estimated task frame provides the directions along/around which the desired
contact force and the desired sensor orientation are prescribed.
Among the task decompositions summarized in~\cref{tab:ledger}, the implemented
controller on observable variables assigns the three translational directions to
force regulation and the two tilt directions to motion regulation.
In particular, prescribing the normal force establishes the normal loading
condition and thus the friction cone within which the commanded tangential force
components must remain.

\looseness=-1 In the following, denote with a star the desired target values
chosen by the user and with a tilde the reference values for the motion
controller.
With reference to~\cref{fig:exp_control_architecture}, the following control
blocks are defined.

\paragraph{Hybrid Target}

The tactile measurement determines the surface normal, but not the twist about
that normal.
Moreover, under~\cref{ass:moment}, no moment estimate is available for
regulating a torque around this direction (see also~\cref{tab:ledger}).
The orientation task is thus prescribed in three angular coordinates, out of
which only the two surface-to-sensor tilt angles of~\cref{eq:output_vector} are
resolved by the contact state.

The target generator issues the hybrid target $(\bm
f?{c}^{\star},\phi^\star,\theta^\star,\psi^\star)$.
The desired contact force $\bm f?{c}^{\star}$ is expressed in the nominal
contact frame, $(\phi^\star,\theta^\star)$ are the desired surface-to-sensor
tilt angles, and $\psi^\star$ is an arbitrary twist angle around the nominal
normal to the surface.
While the current force and tilt components are being reconstructed by the
estimator, the angle $\psi^\star$ has no tactile counterpart.
Anyway, the admittance layer combines all references with $\hat{\bm R}_C$ to
obtain motion references expressed in the world frame.

An admissible tilt target satisfies
\begin{equation}
  \label{eq:tilt_admissible}
  |\phi^\star-\pi|<\frac{\pi}{2}
  \qquad
  |\theta^\star|<\frac{\pi}{2}.
\end{equation}
These ranges are the image of the tilt channels of \cref{eq:h_function} and
define the region in which the sensor approach axis is facing the estimated
plane, i.e.,
\begin{equation}
  \cos \theta^\star \cos \phi^\star <0.
  \label{eq:control_cone}
\end{equation}
Outside this region, the commanded tilt either points the sensor away from the
surface or duplicates an admissible pair.
The range center $(\phi^\star,\theta^\star)=(\pi,0)$ corresponds to the approach
$z$-axis being opposite to the estimated surface normal,
\begin{equation}
  \bm R \bm e_3=-\hat{\bm n}.
  \label{eq:normal_alignment}
\end{equation}
\Cref{eq:tilt_admissible} is a well-posedness condition, and control operation
imposes the tighter restriction already used during acquisition.
Just as the commanded force must remain within the friction cone established by
the normal loading, the commanded tilt must remain within the approach open
half-space~\eqref{eq:control_cone}.

\paragraph{Admittance Layer}

Two complementary admittance channels convert the hybrid targets into
translational and rotational motion references, both mapped to the world frame
through the estimated contact task frame $\hat{\bm R}_C$.

\begin{table}[!t]
  \centering
  \tabstyle
  \caption{Control and estimation parameters used in the closed-loop interaction
    experiment.}
  \label{tab:control_params}
  \begin{tabularx}{\linewidth}{@{}l X c c@{}}
    \toprule
    \textbf{Scope}
    & \textbf{Description}
    & \textbf{Symbol}
    & \textbf{Value} \\
    \midrule

    \multirow{2}{*}{\makecell[l]{Motion\\controller}}
    & Position gain
    & $\bm K_p$
    & $\bm I_3\,\mbox{s}^{-1}$ \\
    & Orientation gain
    & $\bm K_R$
    & $\bm I_3\,\mbox{s}^{-1}$ \\

    \midrule
    \multirow{2}{*}{\makecell[l]{Position\\admittance}}
    & Virtual inertia
    & $\bm M$
    & $\diag(1,1,5)\,\mbox{kg}$ \\
    & Virtual damping
    & $\bm B$
    & $100\diag(1,1,2)\,\mbox{N\,s\,m}^{-1}$ \\

    \midrule
    \multirow{3}{*}{\makecell[l]{Orientation\\admittance}}
    & Virtual inertia
    & $\bm J$
    & $\bm I_3\,\mbox{kg\,m}^2$ \\
    & Virtual stiffness
    & $\bm K$
    & $2\bm I_3\,\mbox{Nm\,rad}^{-1}$ \\
    & Virtual damping
    & $\bm D$
    & $2\sqrt{2}\bm I_3\,\mbox{Nm\,s\,rad}^{-1}$ \\

    \midrule
    \multirow{2}{*}{\makecell[l]{Contact\\detection}}
    & Lower threshold
    & $z*{{lo}}?{{thr}}$
    & $\num{5e-4}\,\mbox{m}$ \\
    & Upper threshold
    & $z*{{up}}?{{thr}}$
    & $\num{7e-4}\,\mbox{m}$ \\

    \midrule
    \multirow{2.6}{*}{\makecell[l]{State\\estimator\\(EKF)}}
    & Initial covariance
    & $\bm P?{{init}}$
    & \makecell[c]{
      $\diag(2,\pi/2,\pi/2,$\\
      $0.1,0.1,0.1)^2$
    } \\
    & Process noise
    & $\bm Q$
    & \makecell[c]{
      $\diag(0.2,\pi/180,\pi/180,$\\
      $0.2,0.2,0.4)^2\Delta t/9$
    } \\

    \bottomrule
  \end{tabularx}
  \tabnote{The estimator state is ordered as in \cref{eq:state_vector}, with
  entries expressed in $[\mbox{m},\mbox{rad},\mbox{rad},
  \mbox{N},\mbox{N},\mbox{N}]$.
    The quantity $\Delta t$ denotes the EKF update period.}
\end{table}

Force regulation is implemented through the position admittance model
\begin{equation}
  \label{eq:pos_admittance}
  \begin{aligned}
    \bm M\dot{\bm\xi}+\bm B\bm\xi
    &=
    \hat{\!\bm f}?{c}
    -
    \bm f?{c}^{\star}\\
    \tilde{\bm v}
    &=
    \hat{\bm R}_C\,\bm\xi\\
    \dot{\tilde{\bm p}}
    &=
    \tilde{\bm v},
  \end{aligned}
\end{equation}
where $\bm M,\bm B\in\mathbb{R}^{3\times3}$ are positive definite, virtual
inertia and damping matrices, respectively.
The admittance state $\bm\xi \in \mathbb{R}^3$ is a velocity resolved in the
estimated contact task frame, whereas $\tilde{\bm v}$ and its integral
$\tilde{\bm p}$ are expressed in the world frame.
The resulting pair $(\tilde{\bm p},\tilde{\bm v})$ defines the translational
reference supplied to the motion controller.

In the rotational channel, the desired angles are first mapped to the target
orientation
\begin{equation}
  \label{eq:ref_orientation}
  \bm R^\star
  =
  \hat{\bm R}_C
  \bm R_z(\psi^\star)
  \bm R_y(\theta^\star)
  \bm R_x(\phi^\star).
\end{equation}
The tilt pair $(\phi^\star,\theta^\star)$ imposes two rotational degrees of
freedom and leaves the twist angle around $\hat{\bm n}$ free.
This remaining degree of freedom is assigned by task
augmentation~\cite[p.~146]{2025-SicVilOriDeL} through an arbitrary choice of
$\psi^\star$, defined from the horizontal $x$-axis of the rotation matrix
in~\eqref{eq:RC_param}.

The target orientation is then processed by the orientation filter
\begin{equation}
  \label{eq:att_admittance}
  \begin{aligned}
    \bm J\dot{\tilde{\bm\omega}}
    +
    \bm D\tilde{\bm\omega}
    +
    \bm K\tilde{\bm e}_R
    &=
    \bm 0\\
    \dot{\tilde{\bm R}}
    &=
    \tilde{\bm R} \,
    \skewm{\tilde{\bm\omega}},
  \end{aligned}
\end{equation}
where
\begin{equation}
  \tilde{\bm e}_R
  =
  \frac{1}{2}
  \veemap{
    \bm R^{\star\top}\tilde{\bm R}
    -
    \tilde{\bm R}^{\top}\bm R^\star
  }.
  \label{eq:orientation_filter_error}
\end{equation}
Here, $\bm J$, $\bm D$, and $\bm K$ are positive definite, virtual inertia,
damping, and stiffness matrices, respectively; moreover, $\skewm{\cdot}$ is the
skew-symmetric operator satisfying $\skewm{\bm u}\bm v=\bm u\times\bm v$ and the
vee operator $\veemap{\cdot}$ denotes its inverse.
Both $\tilde{\bm\omega}$ and $\tilde{\bm e}_R$ are expressed in the tool
reference frame.

Unlike the translational admittance model, the orientation filter is not driven
by the contact force.
Its role is to generate a smooth, kinematically consistent reference
$(\tilde{\bm R},\tilde{\bm\omega})$ so as to track $\bm R^\star$.
The translational channel remains compliant under the force error, while the
rotational channel maintains alignment with the estimated surface.

\paragraph{Motion Controller}

The UR10 manipulator accepts Cartesian velocity commands $(\bm v,\bm\omega)$ for
its tool.
Let the position error and the orientation error be, respectively,
\begin{equation}
  \begin{aligned}
    \bm e_p &=
    \tilde{\bm p} - \bm p\\
    \bm e_R &=
    \frac{1}{2}
    \veemap{
      \bm R^{\top}\tilde{\bm R} - \tilde{\bm R}^{\top}\bm R
    }.
  \end{aligned}
\end{equation}
Given positive definite gain matrices $\bm K_p,\bm K_R\in\mathbb{R}^{3\times3}$,
the commanded Cartesian velocity is
\begin{equation}
  \label{eq:motion_control}
  \bm v
  =
  \tilde{\bm v}+\bm K_p\bm e_p,
  \qquad
  \bm\omega
  =
  \bm{R}^\top \tilde{\bm R} \, \tilde{\bm\omega}+\bm K_R\bm e_R.
\end{equation}
The linear command $\bm v$ and the position error $\bm e_p$ are expressed in the
world frame, whereas the angular command $\bm\omega$ and the orientation error
$\bm e_R$ are expressed in the tool frame.

\Cref{tab:control_params} reports the control gains, detection thresholds, and
estimator parameters used in the experiments of~\cref{sec:results}.

\section{Results}
\label{sec:results}

The acquisition protocol of \cref{sec:dataset} produced $10\,000$ labeled
samples in three batches spanning the indentation and tilt ranges reported in
\cref{tab:exp_params}.
This section first shows the evaluation of the tactile measurement regressor on
the held-out test set and then the assessment of the integrated estimation and
control system in a dynamic contact experiment.

\subsection{Regressor Performance}
\label{subsec:regressor_results}

\begin{figure}[!t]
  \centering
  \includegraphics[
  width=\columnwidth
  ]{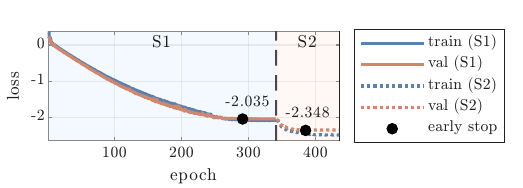}
  \caption{Training and validation losses of the tactile measurement regressor.
    Stage~1 performs homoscedastic pretraining, and Stage~2 performs
    heteroscedastic fine-tuning from the Stage~1 early-stopping checkpoint.}
  \label{fig:training_curves}
\end{figure}

\begin{figure*}[!t]
  \centering

  \subfloat[{Predictions versus labels on the held-out test set.
    Each point represents one test sample, and the dashed line denotes ideal
    agreement.
    The RMSE is reported separately on the top right for each output
    channel.\label{fig:scatter_plots}}]{%
    \begin{minipage}[b]{\textwidth}
      \centering
      \includegraphics[
      width=\textwidth
      ]{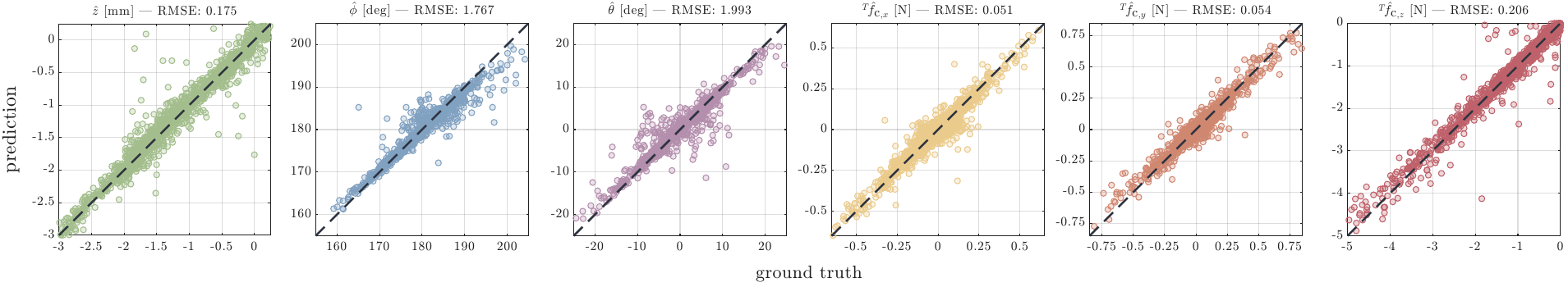}
    \end{minipage}%
  }

  \par\medskip

  \subfloat[{Empirical RMSE versus predicted standard deviation on the held-out
    test set.
    Samples are grouped according to their predicted standard deviation; the
    solid curves report the binned empirical values, and the dashed lines denote
    ideal agreement.\label{fig:calibration_plots}}]{%
    \begin{minipage}[b]{\textwidth}
      \centering
      \includegraphics[
      width=\textwidth
      ]{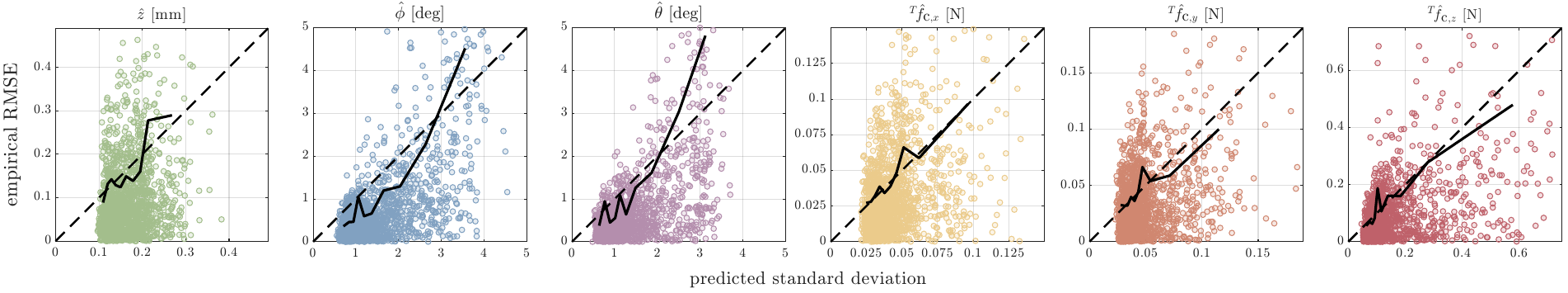}
    \end{minipage}%
  }

  \caption{Performance of the tactile measurement regressor for the six
    observable output variables retained by the contact model.}
  \label{fig:all_plots}
\end{figure*}

\Cref{fig:training_curves} reports the training and validation losses for the
two training stages described in \cref{subsec:training}.
During Stage~1, the validation loss follows the training loss and reaches
$-2.035$ at the early-stopping checkpoint.
Starting from this checkpoint, heteroscedastic fine-tuning reduces the
validation loss to $-2.348$.
Negative values are admissible because \cref{eq:nll_loss} is a Gaussian NLL
evaluated up to an additive constant.
Training from the $10\,000$-sample dataset required $341$ epochs in Stage~1 and
$94$ in Stage~2, approximately $2\,\mbox{h}$ in total on a single NVIDIA
Tesla~T4 GPU.

\Cref{fig:all_plots} reports the performance of the resulting model on the
held-out test set.
The plots of predictions versus labels in \cref{fig:scatter_plots} include the
root mean square error (RMSE) of each output channel.
The indentation RMSE is $0.175\,\mbox{mm}$, while the two surface-to-sensor tilt
angles have RMSEs of $1.767\,\mbox{deg}$ and $1.993\,\mbox{deg}$, respectively.

The indentation error is of the same order as the $0.1\,\mbox{mm}$ repeatability
specified for the manipulator used to construct the geometric labels.
This comparison identifies the accuracy scale of the labeling apparatus; it does
not by itself separate regressor error from label uncertainty.

\looseness=-1 The three force components have RMSEs of $0.051\,\mbox{N}$,
$0.054\,\mbox{N}$, and $0.206\,\mbox{N}$ along the $x$-, $y$-, and $z$-axes of
the tool frame, respectively.
In particular, the RMSEs of the lateral force components are below the nominal
$\num{6.25e-2}\,\mbox{N}$ resolution of the direct force channels of the
labeling sensor.

\looseness=-1 The finer torque resolution of the labeling sensor was also
considered as a possible source of additional supervision for the lateral
components.
Under~\cref{ass:moment}, the ground-truth contact moment satisfies
$\bm\mu?{{gt}}\approx\skewm{\bm l}\bm f?{{gt}}$, where $\bm l$ is the position
of the contact point, expressed in the tool frame, relative to the F/T
measurement origin, so the torque channels provide a redundant measurement of
the labeled force.
Augmenting the loss with the corresponding consistency term $\lVert\skewm{\bm
l}\,{}^T\hat{\!\bm f}?{c} -\bm\mu?{{gt}}\rVert^{2}$ did not, however, produce a
measurable improvement in the accuracy of the lateral force components over
training with \cref{eq:nll_loss} alone, and the term is not part of the reported
model.

\Cref{fig:calibration_plots} compares the predicted standard deviation
$\hat{\sigma}_i$ with the empirical RMSE computed over bins of test samples
sorted by $\hat{\sigma}_i$.
For all six outputs, the empirical error generally increases with the predicted
standard deviation.
The binned curves remain close to the identity over most of the represented
ranges, with larger deviations and variability for the orientation channels.
These results support using $\hat{\bm
W}=\diag(\hat{\sigma}_1^2,\ldots,\hat{\sigma}_6^2)$ as an input-dependent
measurement covariance in the EKF within the tested acquisition domain.

\subsection{Closed-Loop Control Experiment}
\label{subsec:closed_loop_results}

\begin{figure*}[!t]
  \centering

  \subfloat[{Interaction snapshots and corresponding raw tactile images at
    $t=8$, $24$, $40\,\mbox{s}$.
    The rigid board is translated and rotated continuously during the
    experiment.\label{fig:exp_snapshots}}]{%
    \begin{minipage}[b]{\textwidth}
      \centering
      \resizebox{0.32\linewidth}{!}{%
        \includegraphics[
        height=2.4cm,
        ]{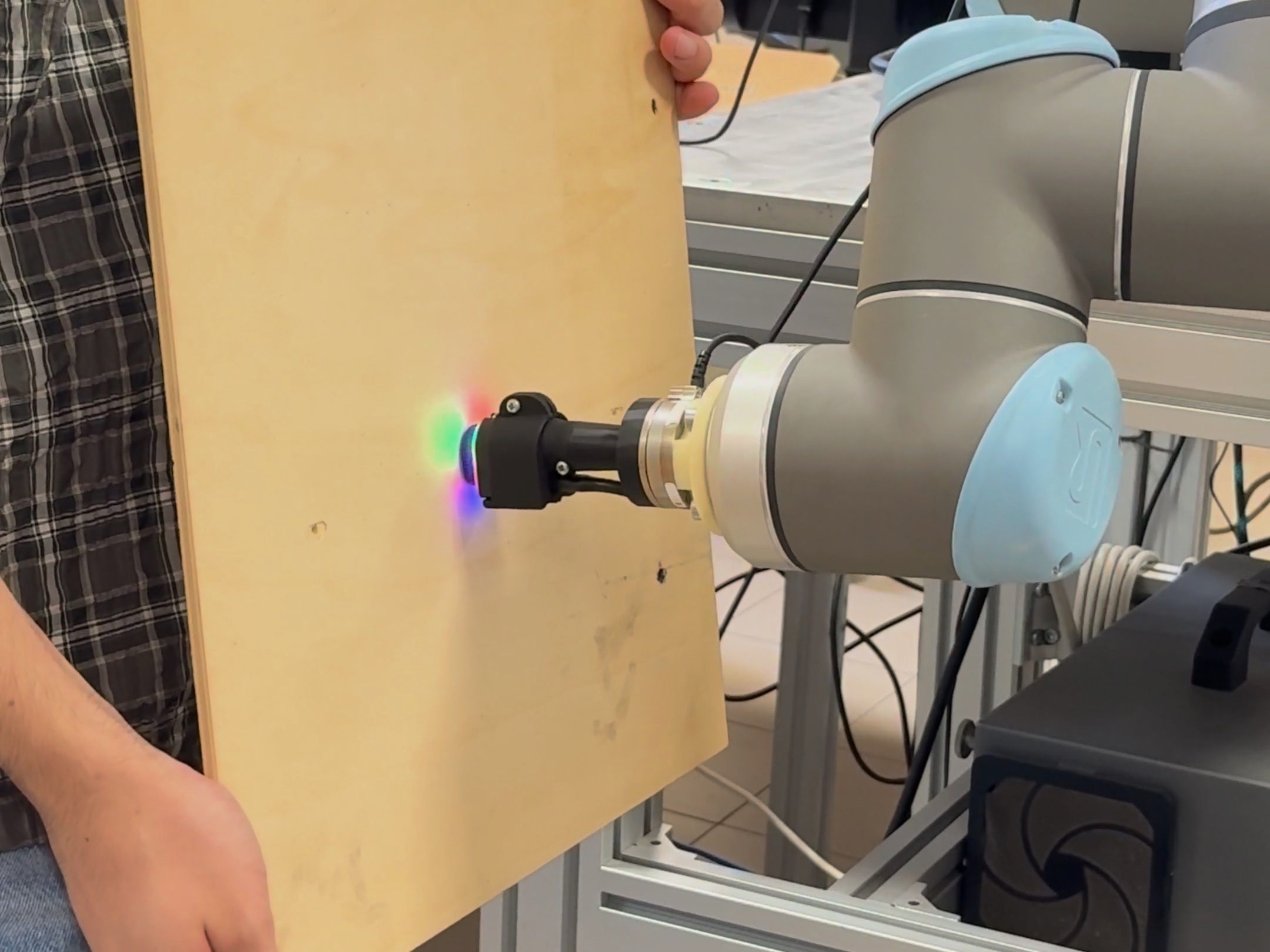}%
        \hspace{0.5mm}%
        \includegraphics[
        height=2.4cm
        ]{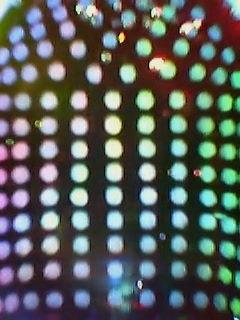}%
      }
      \hspace{1mm}
      \resizebox{0.32\linewidth}{!}{%
        \includegraphics[
        height=2.4cm,
        ]{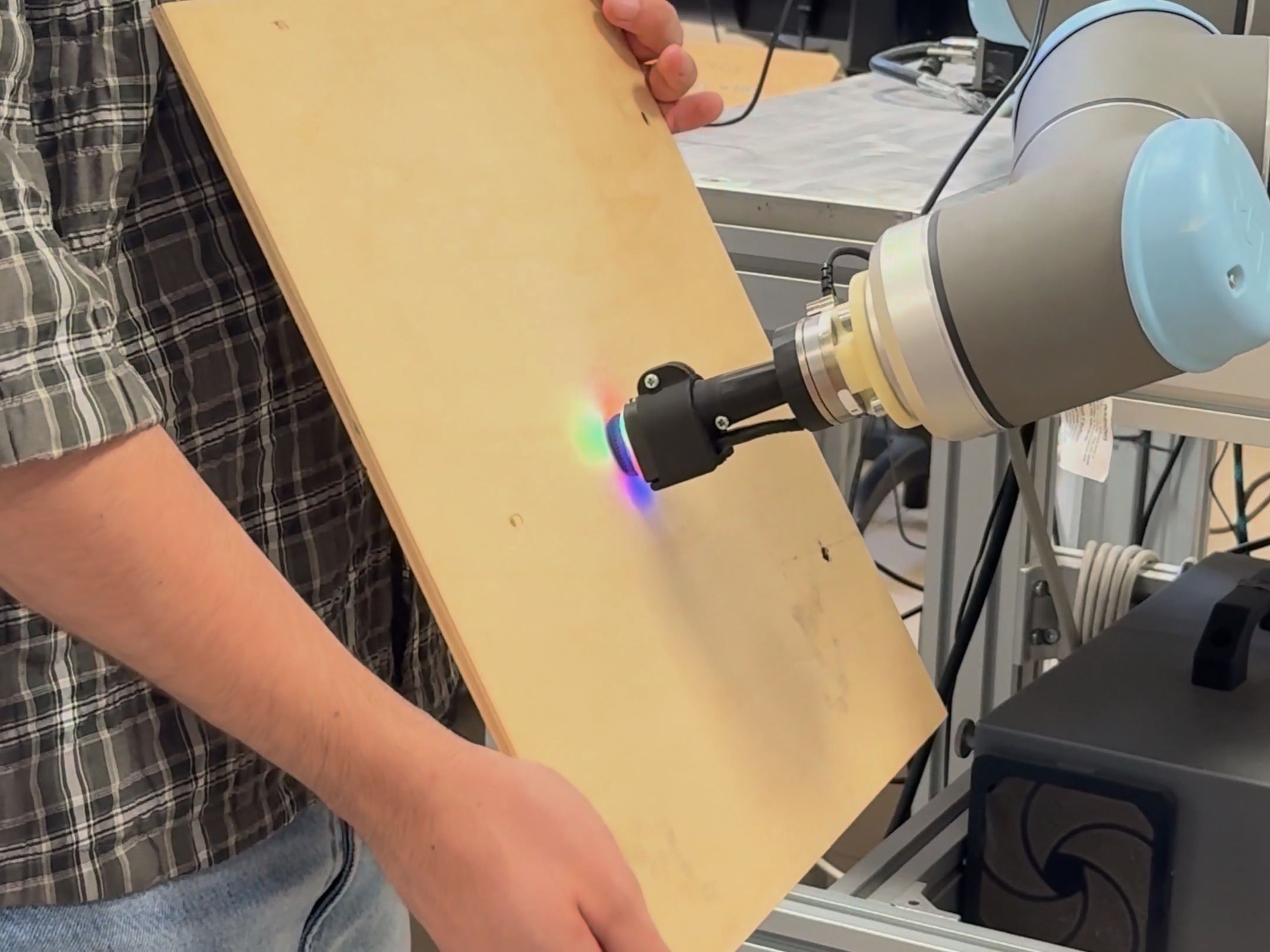}%
        \hspace{0.5mm}%
        \includegraphics[
        height=2.4cm
        ]{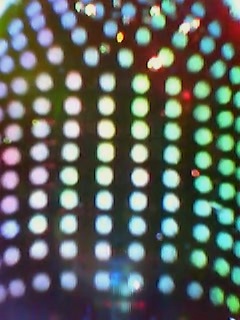}%
      }
      \hspace{1mm}
      \resizebox{0.32\linewidth}{!}{%
        \includegraphics[
        height=2.4cm,
        ]{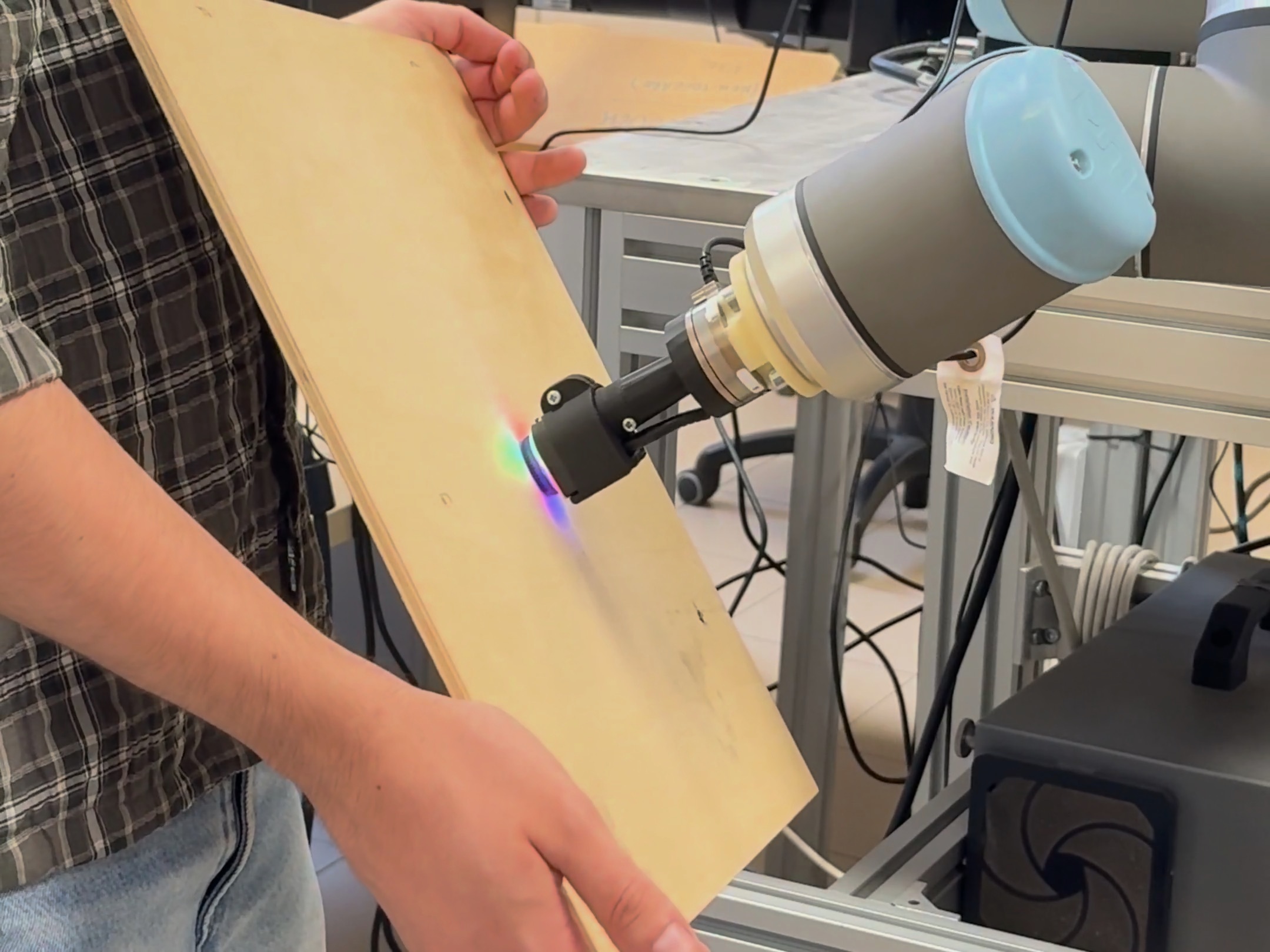}%
        \hspace{0.5mm}%
        \includegraphics[
        height=2.4cm
        ]{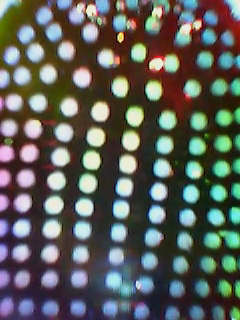}%
      }
    \end{minipage}%
  }

  \subfloat[{Instantaneous tactile force regression ${}^T\hat{\!\bm f}?{c}$ and
    force ${}^T\!\bm f?{{gt}}$ measured offline by the F/T sensor.
    The latter is not used in the control loop.\label{fig:exp_force_gt}}]{%
    \begin{minipage}[b]{0.325\textwidth}
      \centering
      \includegraphics[
      width=\linewidth
      ]{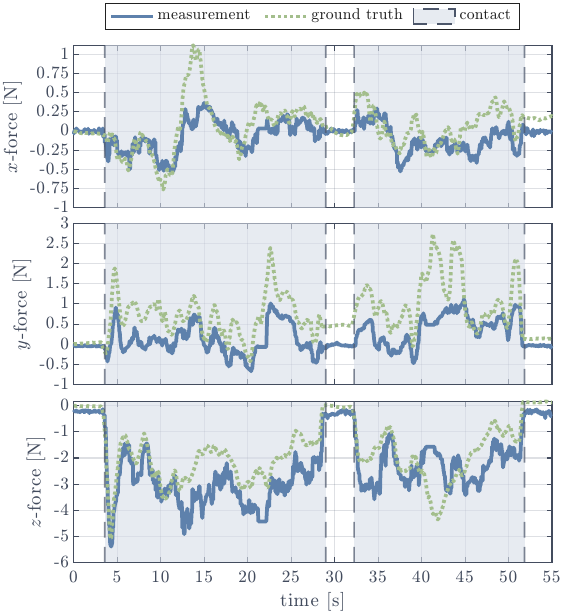}
    \end{minipage}%
  } \,
  \subfloat[{EKF force estimate $\hat{\!\bm f}?{c}$ and prescribed force $\bm
    f?{c}^{\star}$ in the contact task frame.\label{fig:exp_force}}]{%
    \begin{minipage}[b]{0.325\textwidth}
      \centering
      \includegraphics[
      width=\linewidth
      ]{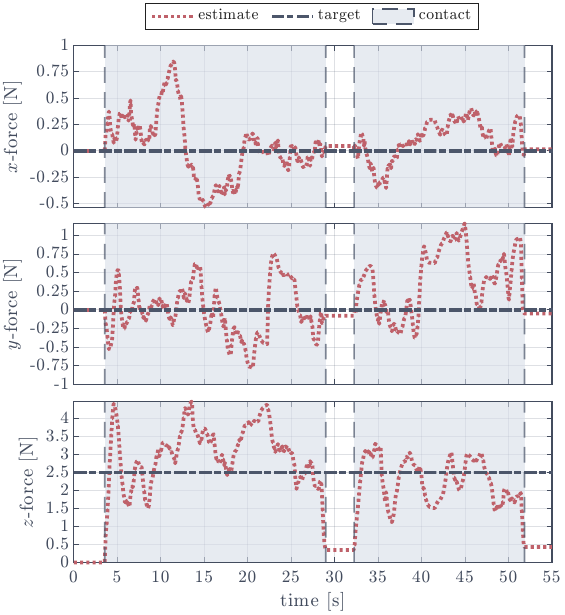}
    \end{minipage}%
  }
  \subfloat[{Tool position $\bm p$ and admittance reference $\tilde{\bm
    p}$.\label{fig:exp_position}}]{%
    \begin{minipage}[b]{0.325\textwidth}
      \centering
      \includegraphics[
      width=\linewidth
      ]{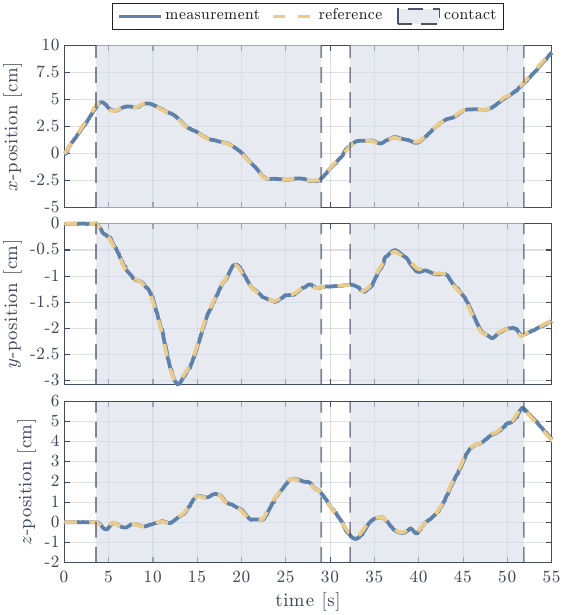}
    \end{minipage}%
  }

  \vspace{-10pt}

  \subfloat[{Sensor-to-surface tilt angles: regressor output $(\hat \phi, \hat
    \theta)$, EKF estimate from $\hat{\bm R}_C^\top \bm{R}$, filtered reference
    from $\hat{\bm R}_C^\top \tilde{\bm{R}}$, and constant target
    $(\phi^\star,\theta^\star)=(\pi, 0)$.\label{fig:exp_cont_rp}}]{%
    \begin{minipage}[b]{0.325\textwidth}
      \centering
      \begin{minipage}[t][6.4cm][b]{\linewidth}
        \centering
        \includegraphics[
        width=\linewidth
        ]{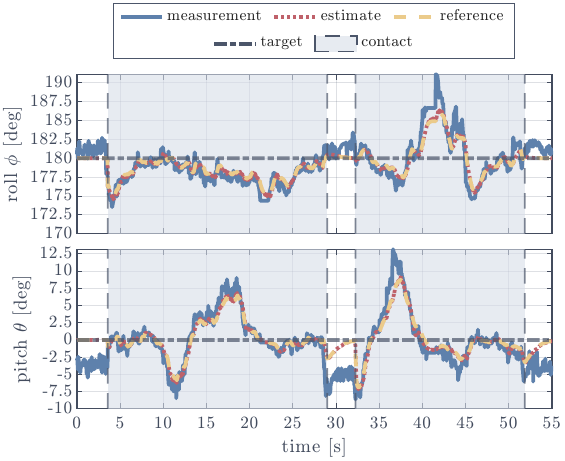}
      \end{minipage}
    \end{minipage}%
  } \quad
  \subfloat[{Current, filtered reference, and target orientations, expressed
    relative to $\bm R?{{init}}$ in~\cref{eq:R_init}: $\bm R?{{init}}^{\top}\bm
    R$, $\bm R?{{init}}^{\top}\tilde{\bm R}$, and $\bm R?{{init}}^{\top}\bm
    R^\star$, respectively.\label{fig:exp_orient}}]{%
    \begin{minipage}[b]{0.325\textwidth}
      \centering
      \begin{minipage}[t][6.4cm][b]{\linewidth}
        \centering
        \includegraphics[
        width=\linewidth
        ]{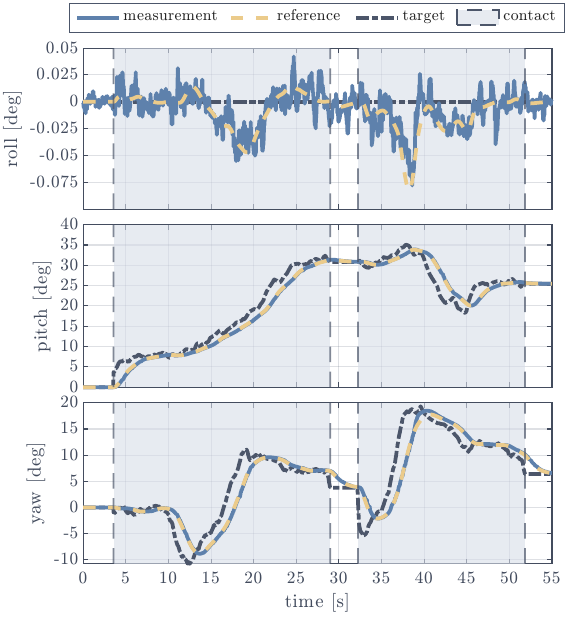}
      \end{minipage}
    \end{minipage}%
  }

  \caption{Experiment of closed-loop tactile force regulation and surface
    alignment against a rigid board moved continuously by a human operator.
    Shaded intervals in the plots over time denote the contact phases detected
    from the regressed indentation depth.}
  \label{fig:exp_demo_plot}
\end{figure*}

The integrated system was evaluated in a dynamic contact experiment.
The tactile regressor, contact state estimator, target generator, and controller
operate at the rates reported in \cref{fig:exp_control_architecture}.
The rate of the Cartesian velocity commands is limited to $20\,\mbox{Hz}$ by the
manipulator interface.

The purpose of this experiment is to establish the feasibility of the complete
feedback architecture, based on touch only, under~\cref{ass:plane,ass:moment}.
It is not intended as an exhaustive characterization across surface materials,
contact geometries, friction conditions, or force references.

The experiment begins out of contact, with the tool at the initial orientation
\begin{equation}
  \bm R?{{init}}
  =
  \bm R_z(\pi/2)\bm R_x(\pi/2).
  \label{eq:R_init}
\end{equation}
During the experiment, an operator continuously translates and rotates a rigid
plywood board, as illustrated in \cref{fig:exp_snapshots}.
The resulting board motion produces an unknown, slowly varying pose of the
contact plane.
The hybrid target is held constant throughout the experiment.
Its force component is the vector in the contact task frame
\begin{equation}
  \bm f?{c}^{\star}
  =
  \begin{pmatrix}
    0 & 0 & 2.5
  \end{pmatrix}^{\top}
  \,\mbox{N},
  \label{eq:experimental_force_reference}
\end{equation}
and its orientation component is the angle triple
\begin{equation}
  (\phi^\star,\theta^\star,\psi^\star)
  =
  (\pi,0,\pi).
  \label{eq:experimental_tilt_reference}
\end{equation}
\looseness=-1 The value $\psi^\star=\pi$ keeps the sensor $x$-axis along the
horizontal line of the estimated contact plane, rotated around the $y$-axis of
that plane by $\theta^\star$.
Thus, all three translational force components are actively regulated: the two
tangential components are regulated to zero, while the normal component is
regulated to $2.5\,\mbox{N}$.
Simultaneously, both observable tilt directions are regulated to the aligned
tilt target of \cref{eq:experimental_tilt_reference}, which aligns the sensor
approach axis with the estimated surface normal.
The experiment therefore jointly exercises tactile measurement, contact
detection, contact state estimation, 3D force regulation, and orientation
tracking.

\videostatement

\begin{table}[!t]
  \centering
  \tabstyle
  \caption{Conditions of the reported closed-loop experiment that differ from
    those of the dataset acquisition campaign.}
  \label{tab:condition_differences}
  \begin{tabularx}{\linewidth}{@{}p{2cm} @{\hspace{0.72cm}} p{2.5cm}
  @{\hspace{0.72cm}} p{3cm}@{}}
    \toprule
    \textbf{Condition} & \textbf{Acquisition} & \textbf{Closed-loop experiment} \\
    \midrule
    Season & Winter & Summer \\
    Contact surface & Polished metal plate & Plywood board \\
    Surface pose & Horizontal, fixed & Near vertical,
    moving \\
    Elastomer skin & Three skins
    & One further skin \\
    \bottomrule
  \end{tabularx}
\end{table}

\looseness=-1 The regressor was used without retraining, fine-tuning, or
recalibration after dataset acquisition, and the experiment was not run under
the acquisition conditions.
\Cref{tab:condition_differences} lists the differences; these factors are
reported only as relevant differences between the training and control
conditions.

The shaded intervals in \crefrange{fig:exp_force_gt}{fig:exp_orient} indicate
the contact phases detected by the Schmitt trigger.
Outside these intervals, the EKF measurement update is disabled because the
tactile output does not represent a valid contact measurement.
During contact, the EKF force estimate $\hat{\!\bm f}?{c}$ is supplied to the
translational admittance controller.

\Cref{fig:exp_force} compares the estimated force with the prescribed reference
in the contact task frame.
The experiment demonstrates simultaneous 3D force regulation: the estimated
normal component approaches its $2.5\,\mbox{N}$ reference, while the two
tangential components remain distributed around their zero references.
Deviations occur principally during changes in the board pose, when the
estimated contact frame, admittance dynamics, and manipulator motion respond to
the externally imposed surface motion.
The signals remain bounded and contact regulation is maintained throughout the
reported experiment.

\looseness=-1 \Cref{fig:exp_force_gt} compares the instantaneous tactile force
prediction ${}^T\hat{\!\bm f}?{c}$ with the force ${}^T\!\bm f?{{gt}}$
reconstructed offline from the F/T sensor, both expressed in the tool frame.
The F/T measurements are used for evaluation only and are not supplied to the
regressor, EKF, target generator, or controller.
The two signals exhibit similar temporal evolution during contact, with the
largest discrepancies occurring during force excursions and contact transients.

\looseness=-1 The principal aggregate results are summarized in
\cref{tab:closed_loop_metrics}.
All force metrics are computed over the samples for which contact is detected.

\begin{table}[!t]
  \centering
  \tabstyle
  \caption{Aggregate results for the control experiment.
  }
  \label{tab:closed_loop_metrics}
  \begin{tabularx}{\linewidth}{@{}L{1.0}C{0.72}C{1.28}@{}}
    \toprule
    \textbf{Quantity} & \textbf{Error} & \textbf{RMSE} \\
    \midrule
    Tactile-F/T
    & $\bm f?{{gt}} - {}^T\hat{\!\bm f}?{c}$
    & $(0.271, 0.755, 1.069)\,\mbox{N}$ \\[2pt]
    Force regulation
    & $\bm f?{c}^{\star} - \hat{\!\bm f}?{c}$
    & $(0.272, 0.466, 0.850)\,\mbox{N}$ \\[2pt]
    Contact tilt wrt normal
    & $( \phi^\star - \hat \phi, \theta^\star- \hat\theta)$
    & $(2.434, 2.992)\,\mbox{deg}$ \\
    \bottomrule
  \end{tabularx}
\end{table}

\looseness=-1 Finally, \cref{fig:exp_orient} reports the measured tool
orientation $\bm R$, the filtered reference $\tilde{\bm R}$, and the target
orientation $\bm R^\star$, all expressed relative to $\bm R?{{init}}$.
As the board orientation varies, the constant orientation target is mapped by
\cref{eq:ref_orientation} into a time-varying $\bm R^\star$ that follows the
contact task frame reconstructed by the EKF.
The filtered reference follows this target, and the measured tool orientation
follows the filtered reference.
This behavior shows that the orientation channel continuously adapts the sensor
alignment to the estimated surface tilt during the experiment.

\begin{figure}[!t]
  \centering
  \resizebox{\columnwidth}{!}{%
    \includegraphics[width=0.48\columnwidth,
    ]{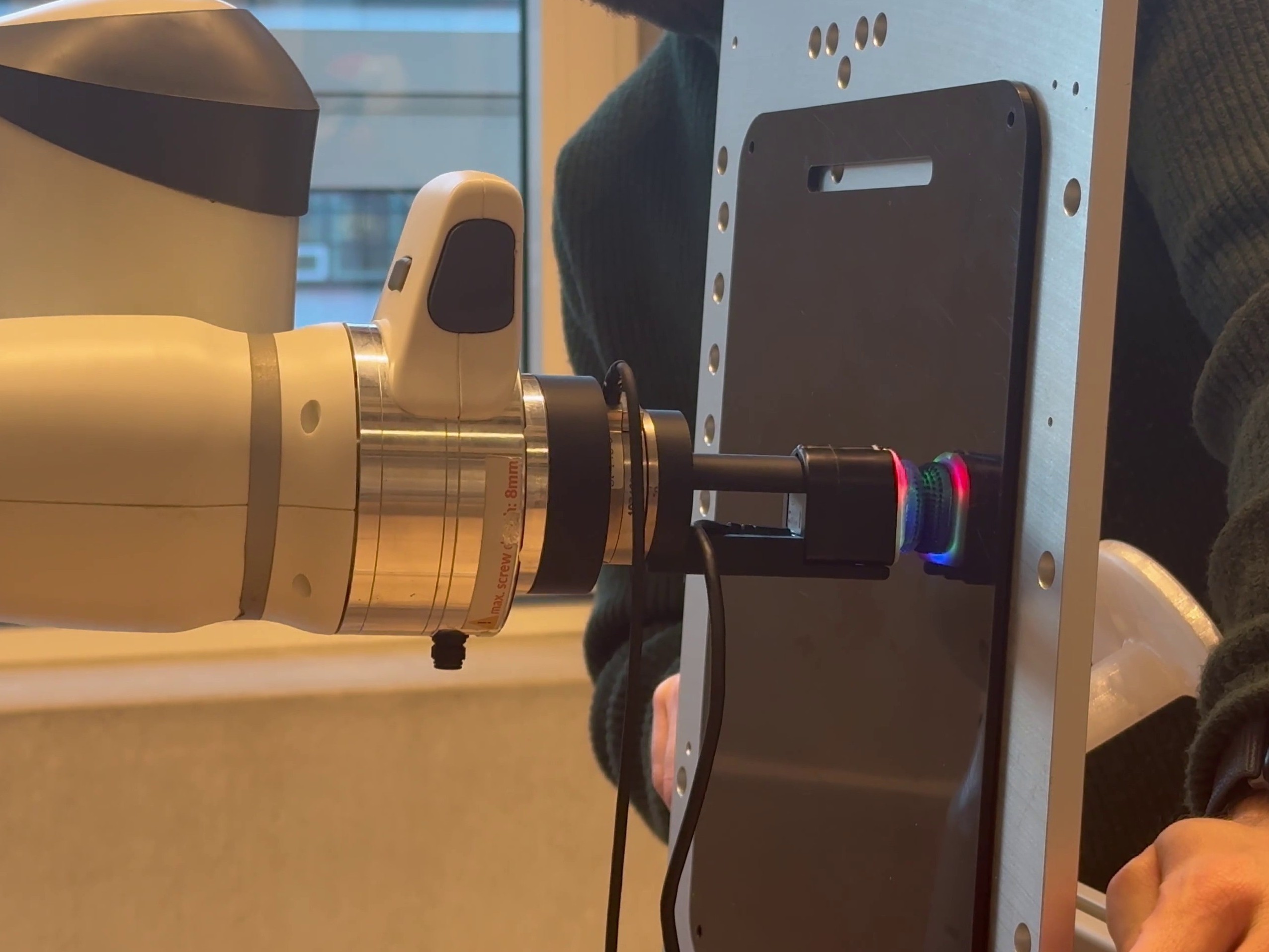}%
    \hspace{3mm}%
    \includegraphics[width=0.48\columnwidth,
    ]{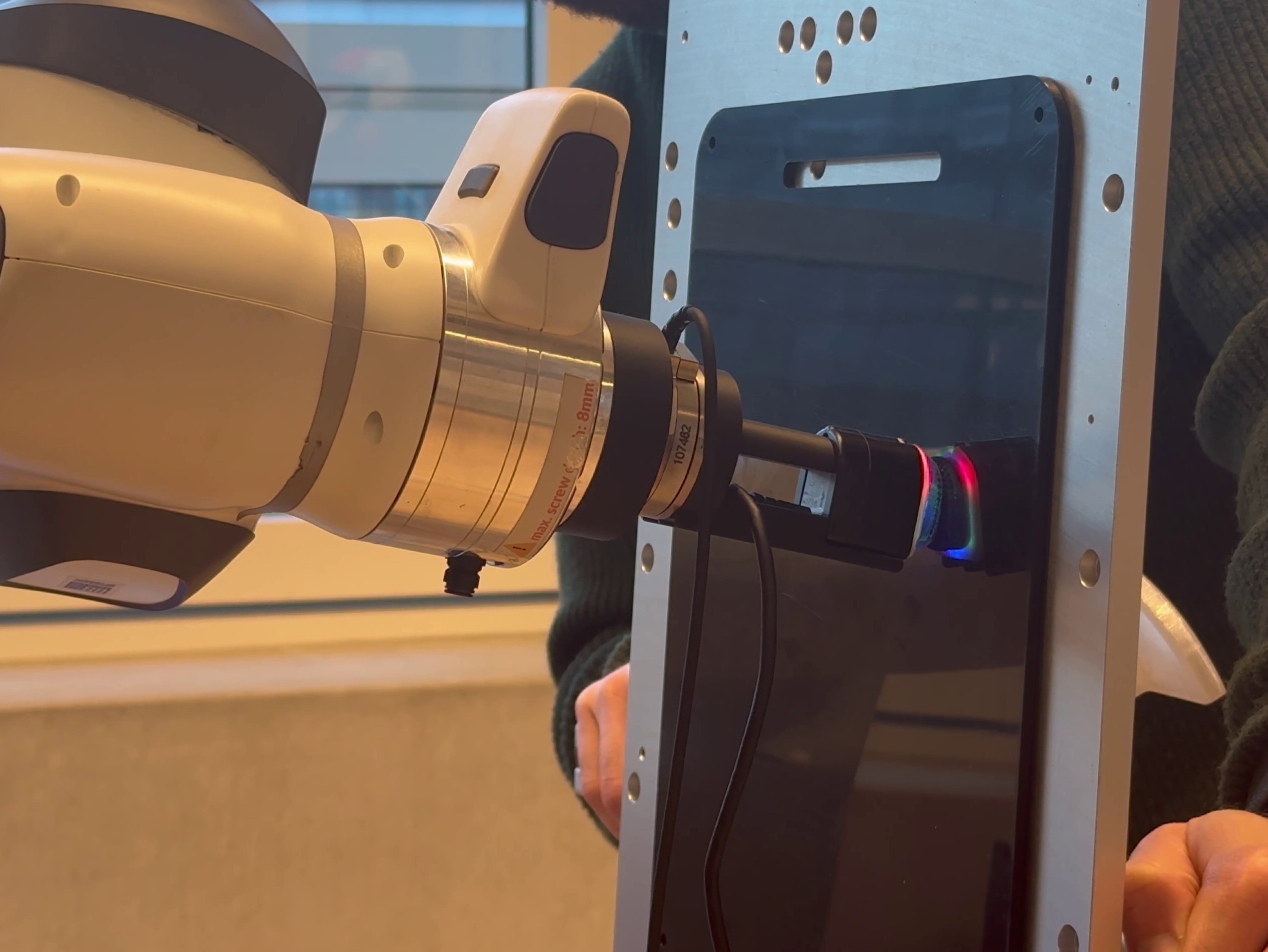}%
  }
  \caption{Two snapshots from trials performed on a different laboratory setup
    than that of~\cref{tab:setup}.
    The regressor of \cref{sec:network} was deployed without any retraining on a
    different manipulator contacting a panel of plastic material.}
  \label{fig:additional_trial}
\end{figure}

\looseness=-1 Between the acquisition campaign and the experiment reported
above, the same regressor was also tested, again without any retraining, on a
different manipulator contacting a vertical plastic panel in a different
laboratory\identifiedonly{ at the University of Twente}
(\cref{fig:additional_trial}).
Those trials used a skin of their own, distinct from the three of the dataset
and from the one of the experiment above.
They addressed different tasks and are not documented here, but rather they are
included as visual evidence that the system was operated outside the setup of
\cref{tab:setup}, with qualitatively comparable behavior.
\additionaltrialvideostatement

\section{Conclusions}
\label{sec:conclusion}

\looseness=-1 Hybrid force/motion control requires knowledge of the interaction
force and of a task frame that suitably defines force- and motion-controlled
directions.
This work addresses their online estimation from soft optical tactile sensing
alone, under the assumptions of locally planar, featureless contact surface and
negligible contact moment.

\looseness=-1 A ResNet-style convolutional regressor maps each tactile image to
an instantaneous measurement including indentation depth, two surface-to-sensor
tilt angles, and 3D contact force,
together with a per-sample uncertainty estimate.
An auxiliary F/T sensor is used offline to construct the ground-truth training
labels and to provide an independent validation reference.
\identifiedonly{The dataset is publicly available (see~\cref{sec:dataset}).}%
\anonymousonly{The dataset will be released upon publication.}

\looseness=-1 The tactile measurement returned by the regressor is fused with
robot proprioceptive data in a contact state estimator based on EKF, producing a
continuously updated estimate of contact task frame and interaction force used
in an admittance-based hybrid force/motion controller.

\looseness=-1 Control experiments have demonstrated contact force regulation
against a flat, rigid board undergoing linear and angular motion imposed by a
human operator.
Within the tested operating conditions, the system was able to keep the
prescribed contact force while adapting the sensor orientation to the estimated
surface normal, using touch provided by the SOT sensor as the only exteroceptive
feedback in the control loop.
The same trained regressor was additionally tested, without any change, on a
different manipulator, touching a different material, in a different laboratory
environment, with qualitatively comparable behavior (see
\cref{fig:additional_trial}).

\looseness=-1 The scope of the obtained results is defined by the contact model
and acquisition domain considered in this work.
The dataset covers rigid, planar, featureless surfaces under moderate
indentation and within the prescribed contact cone.
Thus, we do not claim a priori generalization to larger
shear-to-normal force ratios or to curved, %
textured, or compliant surfaces.
Extending acquisition and measurement models to such contact conditions is an
important \mbox{direction of future work.}

\looseness=-1 The decomposition in \cref{tab:ledger} also identifies directions
that remain available for extending the control formulation.
Rotation about the surface normal is unobservable under the adopted contact
model and is currently fixed through an arbitrary task augmentation.
This remaining degree of freedom could instead be assigned to a secondary
objective, such as avoidance of kinematic singularities or joint limits of the
manipulator.
Furthermore, although tangential contact position is structurally unobservable
on a featureless plane, tangential motion may leave observable signatures
through shear and incipient slip~\cite{2018-JamPesLep}.
A tactile velocity observer could therefore support motion control along
tangential directions when the desired interaction cannot be realized within the
friction cone.

\looseness=-1 We remark that the considered contact task, and in particular the
reconstructed contact variables, is not specific to fixed-base manipulators.
Its integration on a floating-base robotic system (such as an aerial platform)
would require fusion with the corresponding onboard state estimator and an
interaction controller accounting for coupled base and robot dynamics.
Investigating this extension is particularly relevant to operations in which
physical contact is essential and dedicated environment or contact wrench
sensing through a F/T sensor is unavailable.
Finally, beyond force/motion regulation, the same SOT-based estimate of the
contact geometry may constitute an additional local exteroceptive measurement
that could be fused into the platform localization and mapping pipeline,
complementing vision- or lidar-based estimation in visually degraded or
feature-poor environments.
\FloatBarrier

\ifanonymous\else
  \section*{Acknowledgments}
  The authors thank Nicol\`o Botteghi for his valuable feedback on the learning
  methodology.
\fi

\bibliographystyle{IEEEtran}
\bibliography{refs}
\end{document}